\documentclass[letterpaper]{article} 
\usepackage{aaai2027}  
\usepackage[hyphens]{url}  
\usepackage{graphicx} 
\usepackage{natbib}  
\usepackage{caption} 
\usepackage{algorithm}

\usepackage{newfloat}
\usepackage{listings}
\DeclareCaptionStyle{ruled}{labelfont=normalfont,labelsep=colon,strut=off} 
\floatstyle{ruled}
\newfloat{listing}{tb}{lst}{}
\floatname{listing}{Listing}

\usepackage{booktabs}

\usepackage{multirow}
\usepackage{adjustbox}
\usepackage{dsfont}
\usepackage{gensymb}
\usepackage{comment}
\usepackage{upgreek}
\usepackage{algpseudocode}
\usepackage[dvipsnames]{xcolor}
\usepackage{array}
\usepackage{amsmath}
\usepackage{amsfonts}
\usepackage{subcaption}

\newcolumntype{C}[1]{>{\centering\arraybackslash}m{#1}}
\newcolumntype{L}[1]{>{\raggedright\arraybackslash}m{#1}}

\definecolor{HighVPD}{HTML}{C76E00}
\definecolor{LowPrecip}{HTML}{3B82F6}
\definecolor{LowHumidity}{HTML}{00A6D6}
\definecolor{HighTemp}{HTML}{D62728}
\definecolor{NonHighTemp}{HTML}{FFB55A}
\definecolor{NDVIColor}{HTML}{2E8B57}
\definecolor{POPColor}{HTML}{7A52CC}
\definecolor{UNKColor}{HTML}{9E9E9E}

\title{SAE-Xplainers: Rule-Based Feature Interpretation for Extreme Earth Events}
\author{
    Hugo Porta\textsuperscript{\rm 1}\corresponding,
    Emanuele Dalsasso\textsuperscript{\rm 2, \rm 3},
    Chang Xu\textsuperscript{\rm 1},
    Theo Gnassounou\textsuperscript{\rm 1},
    Devis Tuia\textsuperscript{\rm 1}
}
\affiliations {
    \textsuperscript{\rm 1}EPFL, Switzerland\\
    \textsuperscript{\rm 2}Université Grenoble Alpes, France\\
    \textsuperscript{\rm 3}Inria, France\\
    \{hugo.porta, chang.xu, theo.gnassounou, devis.tuia\}@epfl.ch, emanuele.dalsasso@inria.fr
}

\nocopyright
\begin{document}

\maketitle

\begin{abstract}
The emergence of large-scale Weather and Climate (W\&C) datasets offers new opportunities for modeling extreme Earth events (ExEE) and their impacts using deep learning. However, their adoption in operational settings remains limited by the lack of models' interpretability. While for conventional text and image modalities, tools such as Sparse Autoencoders (SAEs) have proven effective for extracting human-understandable concepts, their use for the analysis of ExEE remains challenging due to the nature of W\&C data. To address this, we introduce (i) a geographic location–based modulation of the inputs of SAE to capture the local semantic meaning of environmental patterns, and (ii) an ensemble of rule-based SAE-Xplainers to interpret the resulting high-dimensional features derived from complex, multi-modal environmental predictors. We evaluate our method on three ExEE types: the prediction of \textit{fires}, and the detection of \textit{tropical cyclones} and \textit{atmospheric rivers}. We show that SAE input modulation improves both reconstruction performance and feature utilization, and that our SAE-Xplainers enable faithful interpretation of complex climatic patterns by unfolding them into human-understandable rules that are consistent with the scientific literature, while also supporting the identification of feature absorption.
\end{abstract}

\begin{links}
     \link{Code}{https://github.com/eceo-epfl/SAE-Xplainers}
\end{links}

\section{Introduction}
\label{sec:intro}


Human-induced climate change is causing a widespread increase in the frequency and intensity of ExEE (e.g., wildfires, floods, droughts, storms), with severe consequences for human populations and ecosystems \cite{seneviratne2023weather}. Improving our ability to forecast, detect, and assess the impact of these events is therefore critical for effective preparedness and risk management. The rapid growth of W\&C data and Earth observation imagery \cite{tuia2024artificial} creates new opportunities to address these tasks using deep learning-based models capable of integrating heterogeneous data structures and predictor types \cite{camps2025artificial}.

Nonetheless, the "black box" nature of deep learning models remains a barrier to their adoption in the Earth sciences, especially for extremes, where transparency is essential \cite{camps2025artificial}, prompting the need for eXplainable AI (XAI) methods that interpret model behavior through the location (\textit{where}) \cite{selvaraju2017grad, lundberg2017unified} and semantic meaning (\textit{what}) \cite{kim2018interpretability, koh2020concept} of influential features. In ExEE, most methods aim to answer \textit{where} the model focuses, using feature importance \cite{dikshit2021explainable} or attribution maps \cite{wei2025xai4extremes}. 

\begin{figure}[t]
  \centering
  \includegraphics[width=\linewidth]{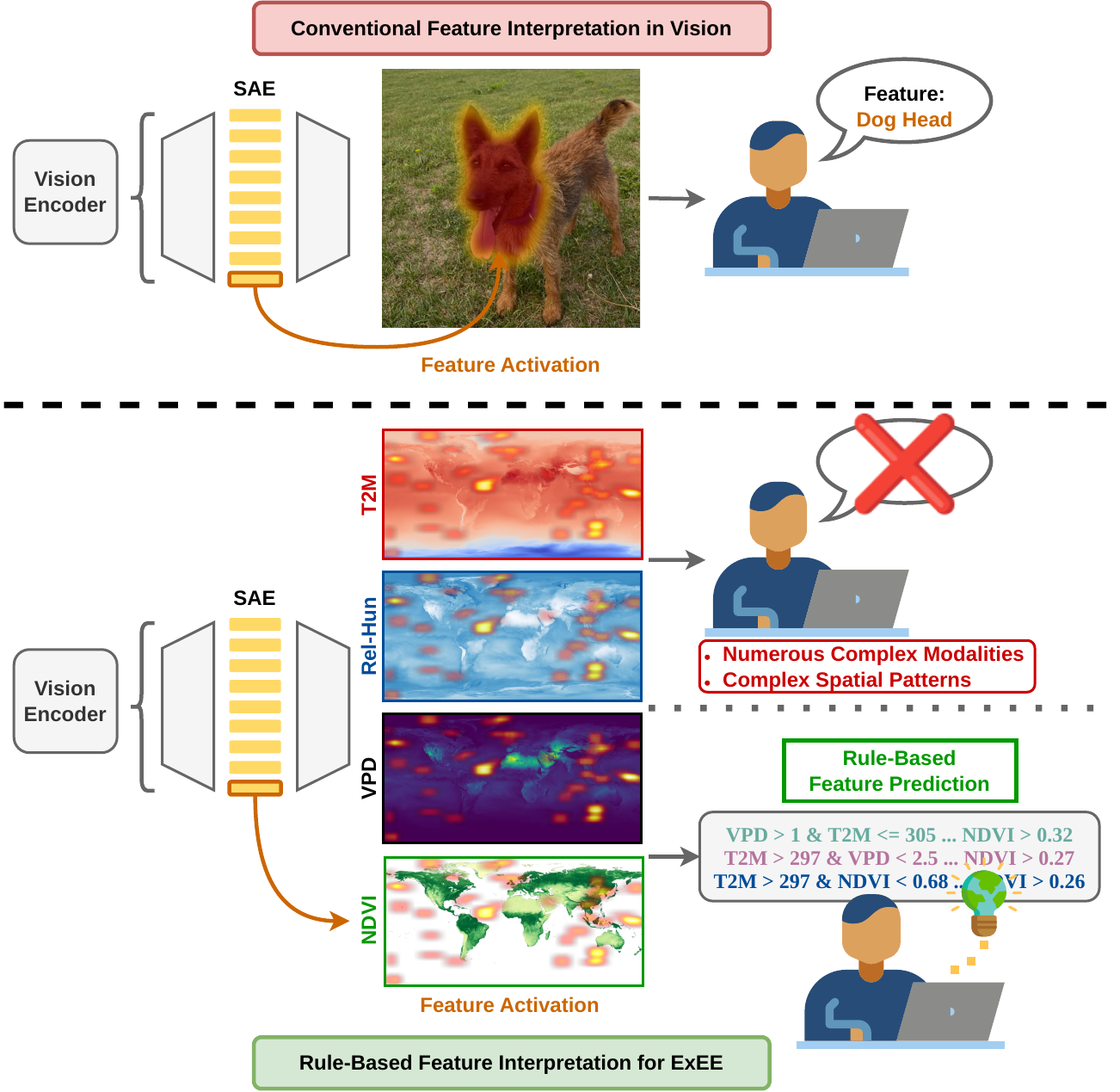}
  \caption{\textbf{(Top):} In computer vision, human users or Vision-Language Models (VLMs) can easily infer the semantic meaning of the SAE features via their activations. \textbf{(Bottom):} In ExEE tasks, the semantic meaning of SAE features cannot be inferred from their activation maps. We propose to learn rule-based models to interpret the features from the input variables (T2M, Rel-Hum, ..., see Table 3 in supplementary material for definition), replacing human or VLM analysis.} 

  \label{fig:mot}
\end{figure}

Alternatively, concept-based methods uncover human-interpretable concepts \cite{kim2018interpretability, koh2020concept} within model representations. Among these, Sparse Autoencoders (SAEs) \cite{huben2024sparse} project polysemantic activations into a sparse latent space to reveal monosemantic "true features" \cite{bricken2023monosemanticity}, with strong effectiveness shown for LLMs \cite{gao2024scaling} and vision models \cite{lim2024sparse}. However, the direct use of SAEs on ExEE data presents limitations in their training and the interpretation of the extracted concepts (see Figure~\ref{fig:mot} and~\ref{fig:motiv-loc}). First, ExEE drivers are often inherently local: similar weather conditions can lead to different outcomes due to regional climate dynamics and environmental context. This represents a challenge for SAEs, which are not designed to learn location-dependent activations (see Figure~\ref{fig:motiv-loc}). Second, existing approaches for the interpretation of SAEs' features, such as LLM autointerpretations \cite{huben2024sparse} and concept activation maps \cite{thasarathan2025universal}, rely on humans or language models. For ExEE, the large number of input modalities and the complex spatio-temporal patterns limit the semantic interpretation of SAEs' features through activation maps in the input space (see Figure~\ref{fig:mot}).

In this paper, we propose an approach for the interpretation of ExEE models that first extracts monosemantic features via a location-aware SAE, and then interprets them using an ensemble of human-understandable rule-based models: SAE-Xplainers. To account for the localized nature of ExEE drivers, we introduce a new geographic location encoder acting as a FiLM adapter \cite{perez2018film}, modulating activations prior to a 
$k$-sparse autoencoder (GeoTopK) \cite{makhzani2013k} to enforce a location-aware projection. We validate our approach on two encoders: ViT \cite{dosovitskiy2020image} (application-agnostic) and ClimaX \cite{nguyen2023climax} (application-specific), across two datasets covering three ExEE types: \textit{fires}, \textit{tropical cyclones}, \textit{atmospheric rivers}, for both forecasting and detection.  
We further show that the monosemanticity of the learned features \cite{bricken2023monosemanticity} enables SAE-Xplainers to faithfully approximate SAE activations (see Figure~\ref{fig:mot}), and that comparing semantic similarity across rules reveals feature absorption \cite{chanin2024absorption}: a sparsity-driven phenomenon producing overly specific, non-atomic features from hierarchical concepts.

In summary, our contributions are as follows: \textbf{(i)} we introduce a geographical location-aware SAE, GeoTopK, to our knowledge, the first domain-conditioned SAE training strategy, which outperforms its location-agnostic counterpart in terms of reconstruction, dead features rate, and on two new application-specific metrics that account for label imbalance and domain shift. \textbf{(ii)} We propose an ensemble of rule-based SAE-Xplainers, which allows faithful interpretation of the SAE features and identification of feature absorption, building a complete SAE interpretability framework for ExEE.

\section{Related Work}

\textbf{Explainable AI in Extreme Earth Events.}
In W\&C, interpretability has mainly been pursued through feature attribution methods addressing \textit{where} the model looks, decomposed into \textit{which predictors, locations,} or \textit{time-steps} most influence the prediction. Shapley additive explanations (SHAP) \cite{lundberg2017unified} and local interpretable model-agnostic explanations (LIME) \cite{ribeiro2016should} have been used to identify the dominant predictors for drought \cite{dikshit2021explainable}, tropical cyclones \cite{cui2025drivers}, wildfires \cite{cilli2022explainable}, and extreme rainfalls \cite{hrast2025expert} while location-sensitive methods such as attention \cite{dikshit2022solving} and Grad-CAM \cite{srinivasan2020machine} have been used to visualize where informative signals arise across space and time. However, these approaches risk increasing user confirmation bias \cite{adebayo2018sanity}, lack global interpretations \cite{van2019global}, raise faithfulness concerns \cite{rudin2019stop}, and face scalability constraints \cite{covert2023learning} for SHAP. Crucially, none address \textit{what} the semantic meaning of influential features is. Disentangling neurons into interpretable concepts remains an open challenge for transparent W\&C models.

\noindent\textbf{Interpretable rule-based models.}
By-design interpretable models, particularly rule-based approaches, have been extensively studied for ExEE applications \cite{rodrigues2022firo, lovo2025tackling}: decision trees offer clear decision paths \cite{breiman2001random}, and methods like Skope-Rules extract compact \texttt{if–then} rules from tree ensembles, though both struggle with complex tasks; Bayesian approaches \cite{wang2017bayesian, letham_interpretable_2015} offer probabilistic rule learning but remain computationally expensive and hard to scale \cite{yang_scalable_2017}. To address this, several works combine deep learning with rule-based structures, including differentiable neural decision trees \cite{kontschieder2015deep}, hypernetworks generating rule parameters from learned representations \cite{yang_hyperlogic_2024}, and Gradient Grafting directly optimizing rules via gradient descent \cite{wang_scalable_2021}, but these methods rely on complex architectures that limit scalability and increase computational cost. Post-hoc methods, which extract rules from a trained model without modifying it, offer an attractive alternative: Anchors, for instance, generate high-precision \texttt{if–then} rules explaining individual predictions \cite{ribeiro_anchors_2018}, but remain only local, describing single-sample predictions. Global feature extraction followed by rule-based interpreters thus appears promising to bridge global understanding with sample-level interpretations.




\noindent\textbf{Sparse Autoencoders for XAI.} 
SAEs are a natural candidate for such global feature extraction as they aim to disentangle the model's polysemantic neurons by projecting them to a sparse high-dimensional space where the dimensions represent monosemantic features \cite{bricken2023monosemanticity}. They have been applied beyond LLMs \cite{huben2024sparse, templeton2024scaling, gao2024scaling} to computer vision, including steering and adaptation analysis of vision-language models \cite{lim2024sparse}, cross-model analysis of vision encoders and pretraining objectives \cite{thasarathan2025universal}, and concept unlearning in diffusion models \cite{bohacek2025uncovering}. Interpreting SAE dictionary vectors as human-understandable concepts typically relies on auxiliary semantic inference via LLM autointerpretations \cite{huben2024sparse, paulo2025automatically}, vision-text similarity measures \cite{lim2024sparse, rao2024discover}, or human inference from activation maps \cite{thasarathan2025universal, fel2025archetypal}.

\begin{figure}[t]
  \centering
  \includegraphics[width=\columnwidth]{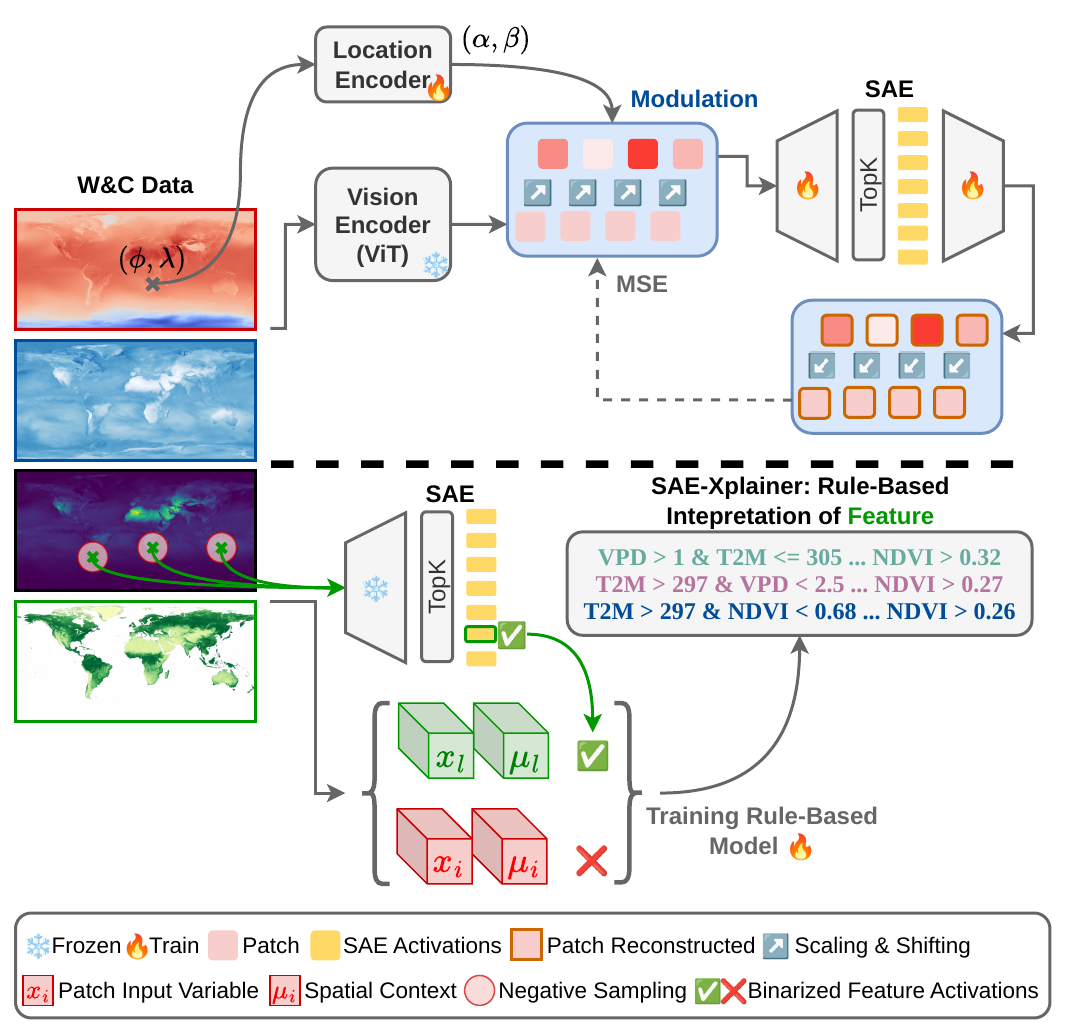}
  \caption{Overview of our framework. \textbf{(Top):} GeoTopK is trained on the modulated patch activations jointly with the location encoder. \textbf{(Bottom):} For a given SAE feature, a SAE-Xplainer is trained via Skope-Rules to predict the binarized activations of selected input patches, where the negatives are sampled adversarially in the neighborhood of positives.}
  \label{fig:method}
\end{figure}

In ExEE, the absence of Weather-Language models prevents the automatic interpretation of SAEs' features, while the complexity of the input modalities and spatial patterns makes human analysis of SAE activations challenging, as shown in Figure~\ref{fig:mot}. We address this with a machine-driven framework for feature interpretations via an ensemble of rule-based SAE-Xplainers. Representing features as rule sets allows us to identify feature absorption \cite{karvonen2025saebench, chanin2024absorption}, a sparsity-driven phenomenon where a hierarchical concept is partially captured by a general feature and partly by more specific ones, causing inconsistent activations and fragmented representations \cite{leask2025sparse}.

\section{Geographic Location-Aware SAE}
\label{sec:geo}
\textbf{SAE Training.} The top panel of Figure~\ref{fig:method} illustrates our approach. We consider a vision transformer-based encoder $f: \mathcal{X} \rightarrow \mathcal{Y}$, with $f^{(l)}: \mathcal{X} \rightarrow \mathbb{R}^{T \times D}$ the mapping from the input to the sequence $T$ of tokens activations at layer $(l)$. For an input sample $x \in \mathcal{X}$, we define the matrix $H^{(l)} = f^{(l)}(x) \in  \mathbb{R}^{T \times D}$ as the latent activations of the model for layer $l$. For a specific patch (token) index  $t \in \{1, ..., T\}$, we denote $x_t$ the corresponding input patch and $h_t^{(l)} = H_{t,:}^{(l)} \in \mathbb{R}^{D}$ its latent activations at layer $l$. SAEs aim to learn a sparse and overcomplete decomposition of the token activation $h_t^{(l)}$ into a high-dimensional space of size $N$. Experimentally, we focus on the last residual stream output of the transformer ($h_t$) as input to the SAE. $h_t$ is passed through a $k-$sparse autoencoder to directly control the number of active features $\mathcal{Z}$ via the TopK function, which replaces the ReLU activation and the $L_1$ sparsity penalty. The encoder and decoder operations are defined by:
\begin{equation}
\begin{aligned}
    z &= \text{TopK}(W_{enc}h_t + b_{enc})\;, \\
    \hat{h}_t &= W_{dec}z\;,
\end{aligned}
\label{eq:sae}
\end{equation}
with $W_{enc} \in \mathbb{R}^{N \times D}$, $b_{enc} \in \mathbb{R}^N$, and $W_{dec} \in \mathbb{R}^{D \times N}$. Following \cite{gao2024scaling}, we also apply ReLU before TopK to ensure non-negative activations in our code. The decoder matrix $W_{dec}$ consists of $N$ column vectors, where each vector $d_i$ represents a monosemantic feature learned by the model. The vector $z \in \mathbb{R}^N$ represents the feature activations and $\|z\|_0 = k$.

SAEs are trained to minimize the mean squared error (MSE) as a reconstruction objective. Architectures like $k-$sparse autoencoders also require the use of an auxiliary loss $\mathcal{L}_{\text{aux}}$ to revive inactive features \cite{gao2024scaling}, its computation is detailed in the supplementary material:

\begin{equation}
    \mathcal{L}_{\text{SAE}} = \|h_t-\hat{h}_t\|^2_2 + \gamma \mathcal{L}_{\text{aux}}\;.
\label{eq:loss}
\end{equation}

\begin{figure}[t]
  \centering
  \includegraphics[width=\columnwidth]{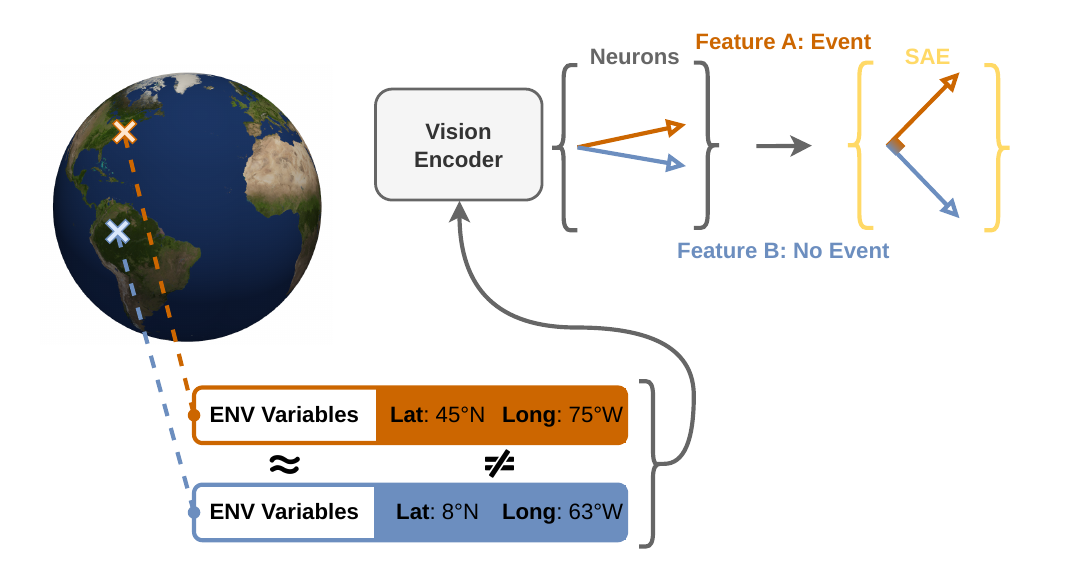}
  \caption{Similar environmental conditions (ENV) can lead to different ExEE probabilities due to local effects, requiring models to learn location-dependent features.}
  \label{fig:motiv-loc}
\end{figure}

\noindent\textbf{Geographic Location Modulation.} To capture the local nature of ExEE, we enforce geographic location information into the activations of the models, enhancing the SAE ability to disentangle location-specific characteristics of the input predictors, as illustrated in Figure~\ref{fig:motiv-loc}. Given the geographic location set $\mathcal{G}$, we learn a geographic location encoder $\Psi: \mathcal{G} \subset  \mathbb{R}^{2} \rightarrow \mathbb{R}^{2D}$, 
which maps latitude $\phi \in [-90, 90]$ and longitude $\lambda \in [-180, 180]$ of an input patch $x_t$ to the modulation parameters $(\alpha, \beta) \in \mathbb{R}^{2D}$:

\begin{equation}
\begin{aligned}
    \Psi(\phi, \lambda) &= \text{MLP}([\sin(\phi), \cos(\phi), \sin(\lambda), \cos(\lambda)])\;, \\
    \Psi(\phi, \lambda) &= (\alpha, \beta)\;.
\end{aligned}
\end{equation}
The modulation parameters are used to perform feature-wise affine modulation of the model activation $h_t$, acting as FiLM adapters. This approach is used in normalization layers, where shift-only transformations cannot adjust the importance of neurons in the model activations, and scalar modulations can lack expressiveness. Moreover, feature-wise modulation provides expressiveness without the computational overhead and neuron mixing of full linear layers. We further apply the $\tanh$ nonlinearity to constrain the magnitude of both scaling and shifting, ensuring stable conditioning and preventing the SAE from trivially encoding only geographic location, resulting in:
\begin{equation}
    h_{t, \text{mod}} = (1 + \tanh(\alpha)) \odot h_t + \tanh(\beta)\;.
\end{equation}
The vector $h_{t,\text{mod}}$ is then passed to the SAE in Equation~\ref{eq:sae} in place of $h_t$, and the affine modulation is inverted before the reconstruction, as shown in Figure~\ref{fig:method}. Ablation results on the geographic location modulation parameterization and comparison with other methods \cite{eddin2023location} are provided in the supplementary material Table 6. 

\noindent\textbf{Evaluation.} ExEE tasks are highly imbalanced and span diverse climatic regimes across the globe, requiring models to generalize across domains. To account for these challenges, we extend the standard metrics used in SAE evaluation: $\text{R}^2$, MSE, and ratio of dead features. 

First, we analyze reconstruction fidelity by conditioning the computation of $\text{R}^2$ on extreme events, which we denote as the \textbf{$\text{R}^2_\text{event}$}$ = \text{R}^2\!(h_t, \hat{h_t})\big|_{\mathcal{I}_{\text{Event}}}$,
where $\mathcal{I}_{\text{event}}$ are the indices of input patch samples in $\mathcal{X}$ belonging to extreme events. This metric is macro-averaged across event classes and provides an assessment of reconstruction fidelity on the target classes despite the severe class imbalance.

Second, we condition the computation of $\text{R}^2$ across predefined latitude bands and report the \textbf{$\text{R}^2_{\text{worst}}$}$=
\min_{b \in \mathcal{B}}\text{R}^2\!(h_t,\hat{h_t}\mid \phi \in b)$, with latitude intervals of $20^\circ$  $\mathcal{B} = \{[-90,-70), [-70,-50), \dots, [70,90]\}$. This metric captures the worst-case reconstruction performance of the SAE across geographically diverse domains, providing a lower bound on reconstruction fidelity at the global scale. This echoes the worst-group accuracy metric, often used in applications with domain shift to assess model robustness.

\section{Rule-based SAE-Xplainers}

\textbf{Training the SAE-Xplainers.} First, we rewrite Equation~\ref{eq:sae} using the feature vectors $d_i \in \mathbb{R}^{D}, \forall i \in [1, N]$:

\begin{equation}
    \hat{h}_t = W_{dec}z = \sum_{i=1}^{N}z_i d_i\;,
\end{equation}
with $z_i$ the activation of the feature vector $d_i$. An emergent property of SAEs' high-dimensional and sparse latent space is the monosemanticity of its features \cite{elhage2022superposition}. As a consequence, we hypothesize that we can learn for each feature vector $d_i$ a surrogate interpretable model $g_i: \mathcal{X} \rightarrow \{0, 1\}$ that faithfully approximates the mapping from the multi-modal input patch to its binarized activations $\tilde{z}_i(x_t) = \mathds{1}[z_i(x_t) > 0]$, as illustrated in the bottom panel of Figure~\ref{fig:method}. The set of $N$ surrogate models constitutes our ensemble of SAE-Xplainers. Due to the high sparsity of SAE features, their activations follow a zero-inflated distribution, with most samples corresponding to zero activations. Therefore, interpretability depends more on accurate distinctions between activating and non-activating samples, rather than on predicting precise activation magnitudes \cite{karvonen2025saebench}. 
A model $g_i$ is defined as a finite set of $m_i$ rules $\mathcal{R}_i = \{r_{i, 1}, ..., r_{i, m_i}\}$, where each rule $r_{i, j}: \mathcal{X} \rightarrow \{0, 1\}$ is a conjunction of threshold conditions on the input variables with a normalized weight $w_j$:
\begin{equation}
    \begin{aligned}
    g_i(x_t) = \mathds{1}[\sum_{j=1}^{m_i} w_j r_{i,j}(x_t) > 0.5]\;.
    \end{aligned}
\label{eq:sur}
\end{equation}
The rule-set surrogate models are trained using the Skope-Rules algorithm under multiple parameterizations to identify the proper level of complexity (measured as the number of rule conditions per rule set) for predicting feature activations. 
To further improve interpretation quality, we extend the top-random balanced sampling strategy introduced in LLM autointerpretability \cite{huben2024sparse, paulo2025automatically} with a novel adversarial geographic negative sampling scheme: for each surrogate model $g_i$, negative samples are drawn within a geographic radius around the positive activation set $\mathcal{I}_i = \{x_t \mid z_i(x_t) > 0, \: x \in \mathcal{X}\}$, ensuring better rule-based interpretations and preventing location-based discrimination (see Figure~\ref{fig:method}). The adversarial negative set is defined as:

\begin{equation}
\mathcal{I}_{\text{neg}, i}
=
\{
x_t \mid
z_i(x_t) = 0
\; \wedge \; d^*(x_t)
\leq \sigma, \; x \in \mathcal{X}
\}\;,
\end{equation}
with $d^*(x_t) = \min_{x_t^+ \in \mathcal{I}_i} d_{\mathbb{S}^2}\big(c(x_t), c(x_t^+)\big)$, $d_{\mathbb{S}^2}(., .)$ the geodesic distance on the Earth as a sphere, and $\sigma$ the sampling radius. A comparison of the interpretations across sampling strategies is provided in the supplementary material.

Our approach is similar to Anchors \cite{ribeiro_anchors_2018}, as it extracts rule-based interpretations of the model behavior. However, while Anchors generate local rules explaining individual predictions in the input space via perturbations, our SAE-Xplainers learn surrogate rules on sparse latent activations to provide scalable, global interpretations of learned features in the representation space.

\noindent\textbf{Integrating Spatial Context.} The surrogate model formulation (Equation~\ref{eq:sur}) does not account for the influence of spatial context on feature activations. However, the importance of spatial interactions between patches on the learned features can vary depending on the event types, as shown in Figure~\ref{fig:xplain-quant}a. To capture this effect, we extend Equation~\ref{eq:sur} so that the surrogate models account for statistics computed over the set of most attended patches for a given input patch $x_t$, where attention scores are aggregated across all layers of the transformer model $f$ up to layer $l$: $A^{l}(x_t) \in \mathbb{R}^{T}$. We denote this new set of most attended patches as $\mathcal{N}_n(x_t) = \text{TopK}_n(A^{l}(x_t) \mid t \in \{1, ..., T\})$. The spatially aware surrogate model $g_i$ becomes:

\begin{equation}
g_i(x_t)=\mathds{1}[\sum_{j=1}^m w_j r_{i,j}(x_t, \mu_t^{(n)}, m_t^{(n)}, \ell_t^{(n)}) > 0.5]\;,
\label{eq:spa-att}
\end{equation}

with $\mu_t^{(n)}=\underset{s \in \mathcal{N}_n(x_t)}{\operatorname{mean}}x_s$, $m_t^{(n)}=\underset{s \in \mathcal{N}_n(x_t)}{\operatorname{max}}x_s$, and $\ell_t^{(n)} = \underset{s \in \mathcal{N}_n(x_t)}{\operatorname{min}}x_s$.\\

\noindent\textbf{Evaluation.}
The rule-based interpretations from our ensemble of SAE-Xplainers are first evaluated on their ability to faithfully predict the activations of the SAE features, as shown in Figure~\ref{fig:xplain-quant}a. Nonetheless, this metric highly depends on the sampling strategy used for building the training and test sets and is not enough to guarantee the quality of feature interpretations. To this end, we investigate the potential of SAE-Xplainers to identify feature absorption at scale across all SAE features. Feature absorption \cite{chanin2024absorption} is a sparsity-induced phenomenon in SAEs, where hierarchical concepts are not represented as separate atomic features, but instead a specific feature may absorb part of a more general one, leading to fragmented, overly specific, and less semantically coherent representations. If our rule-based interpreters are able to capture feature absorption, then the learned rules are consistent with the SAE feature activations.

We use our SAE-Xplainers to identify groups of semantically related features. First, we filter the set of SAE features into groups of connected components, where edges between two features are defined by a Jaccard distance below a given threshold, computed over their rule sets using the Hungarian algorithm for optimal assignment, retaining only the $M$ closest pairs. The identified set of $P$ partitions $\{\mathcal{K}_j\}_{j=1}^{P}$, with $\mathcal{K}_j \subseteq \{1, \dots, N\}$, define candidate feature groups sharing highly similar rule interpretations. Within each group $\mathcal{K}_j$, we measure feature co-occurrence based on their activation sets 
$\mathcal{I}_i = \{x_t \mid z_i(x_t) > 0, \: x \in \mathcal{X}\}$. For a feature $d_i \in \mathcal{K}_j$, the total group co-occurrence ratio is

\begin{equation}
\rho_i=\frac{\sum_{d_l \in \mathcal{K}_j \setminus \{d_i\}}|\mathcal{I}_i \cap \mathcal{I}_l|}{|\mathcal{I}_i|}\;.
\end{equation}
A feature $d_i$ is considered absorbed if $\rho_i \geq \tau$, where $\tau$ is a fixed threshold. We then define the \textbf{co-occurrence score} as the ratio of groups containing at least one absorbed feature.

Finally, we report the SAEBench \textbf{absorption score} \cite{karvonen2025saebench} within co-occurring subgroups, using as main feature $d_{i^\star} = \max_{d_i \in \mathcal{K}_j} \rho_i$, and defining the prediction task with its rule-based surrogate model $g_{i^\star}$. This evaluates whether the rules learned by the SAE-Xplainers can identify co-occurring features that encode absorbed concepts.

\section{SAE Performance Analysis}

\textbf{Experimental Setup.} We test our approach on two global datasets and tasks. First is an 8-day horizon dense forecasting of \textit{fires} using the SeasFire dataset \cite{karasante2025seasfire}, comprising a set of 14 variables (including 10 W\&C variables, a biosphere variable, a socioeconomic variable, and the geographic coordinates). Second is the detection of \textit{tropical cyclones} and \textit{atmospheric rivers} from ClimateNet \cite{prabhat2020climatenet} on a set of 6 variables (including 4 W\&C variables, and the geographic coordinates). For both tasks, we fine-tuned a W\&C foundational model: ClimaX \cite{nguyen2023climax}, and a generic vision encoder: ViT \cite{dosovitskiy2020image}. Both use a patch size of $4 \times 4$. As a result, our input patch token $x_t$ and its corresponding model activations $h_t$ represent $4 \times 4$ pixels at SeasFire and ClimateNet native resolution of $\sim 0.25^{\degree}$.
For each model and dataset, we train the TopK SAE baseline \cite{makhzani2013k} and our geographic location-aware GeoTopK. All results use $k=4$ and SAE dimension $N=4 \cdot D=4096$, favoring high sparsity to strengthen interpretability for human experts. Training and dataset details are provided in supplementary material.

\begin{table}[t]
\centering
\begin{subtable}{\linewidth}
\centering
\small
\begin{tabular}{l cc cc}
\toprule
\multirow{2}{*}{\textbf{Metric}}
& \multicolumn{2}{c}{\textbf{ClimaX}}
& \multicolumn{2}{c}{\textbf{ViT}} \\
& \textbf{TopK} & \textbf{GeoTopK}
& \textbf{TopK} & \textbf{GeoTopK} \\
\cmidrule(lr){1-1}
\cmidrule(lr){2-3} \cmidrule(lr){4-5}
$\text{R}^2$ $\uparrow$
& 0.927 & \textbf{0.937} & 0.825 & \textbf{0.840} \\
$\text{R}^2_{\text{worst}}$ $\uparrow$
& 0.881 & \textbf{0.901} & 0.767 & \textbf{0.786} \\
$\text{R}^2_{\text{event}}$ $\uparrow$
& 0.801 & \textbf{0.807} & \textbf{0.651} & 0.633 \\
MSE $\downarrow$
& 1.84e-3 & \textbf{1.59e-3} & 1.77e-1 & \textbf{1.63e-1} \\
Dead Feat. $\downarrow$
& 0.159 & \textbf{0.002} & 0.021 & \textbf{0.000} \\
\bottomrule
\end{tabular}
\caption{Results on SeasFire.}
\label{tab:seasfire-result}
\end{subtable}

\begin{subtable}{\linewidth}
\centering
\small
\begin{tabular}{l cc cc}
\toprule
\multirow{2}{*}{\textbf{Metric}}
& \multicolumn{2}{c}{\textbf{ClimaX}}
& \multicolumn{2}{c}{\textbf{ViT}} \\
& \textbf{TopK} & \textbf{GeoTopK}
& \textbf{TopK} & \textbf{GeoTopK} \\
\cmidrule(lr){1-1}
\cmidrule(lr){2-3} \cmidrule(lr){4-5}
$\text{R}^2$ $\uparrow$
& 0.943 & \textbf{0.953} & 0.987 & \textbf{0.993} \\
$\text{R}^2_{\text{worst}}$ $\uparrow$
& 0.910 & \textbf{0.924} & \underline{0.983} & \underline{0.988} \\
$\text{R}^2_{\text{event}}$ $\uparrow$
& 0.880 & \textbf{0.893} & 0.974 & \textbf{0.980} \\
MSE $\downarrow$
& 1.01e-3 & \textbf{0.84e-3} & 12.85e-3 & \textbf{7.12e-3} \\
Dead Feat. $\downarrow$
& 0.513 & \textbf{0.391} & 0.811 & \textbf{0.442} \\
\bottomrule
\end{tabular}
\caption{Results on ClimateNet.}
\label{tab:climatenet-result}
\end{subtable}

\caption{Quantitative Comparison of TopK and GeoTopK on ClimaX and ViT last residual stream, fine-tuned on SeasFire and ClimateNet, respectively. \textbf{Bold} means best score, \underline{underline} means similar score (difference is $\leq 0.005$).}
\label{tab:main-result}
\end{table}

\noindent\textbf{GeoTopK Performance.} Table~\ref{tab:main-result} shows the contribution of our geographic location-aware SAE: GeoTopK on SeasFire and ClimateNet test sets. We observe that across all reconstruction metrics and almost all setups, our GeoTopK outperforms the TopK baseline, showing the importance of geographic location modulation. Moreover, GeoTopK highly reduces the number of dead features, which is a key aspect for SAEs that favors monosemanticity. This is more pronounced in the ClimateNet dataset, where standard TopK architectures fail to maintain even a 50\% feature activation rate, despite the implementation of auxiliary losses and feature resampling techniques. By integrating geographic location modulation, GeoTopK effectively prevents the emergence of polysemantic features typically caused by neural superposition. The impact of scaling the latent dimension size of the SAE is shown in supplementary material Figure 7. 

\section{Rule-based Feature Interpretations}
\label{sec:rules}

\begin{figure*}[t]
    \centering
    \includegraphics[width=0.9\textwidth]{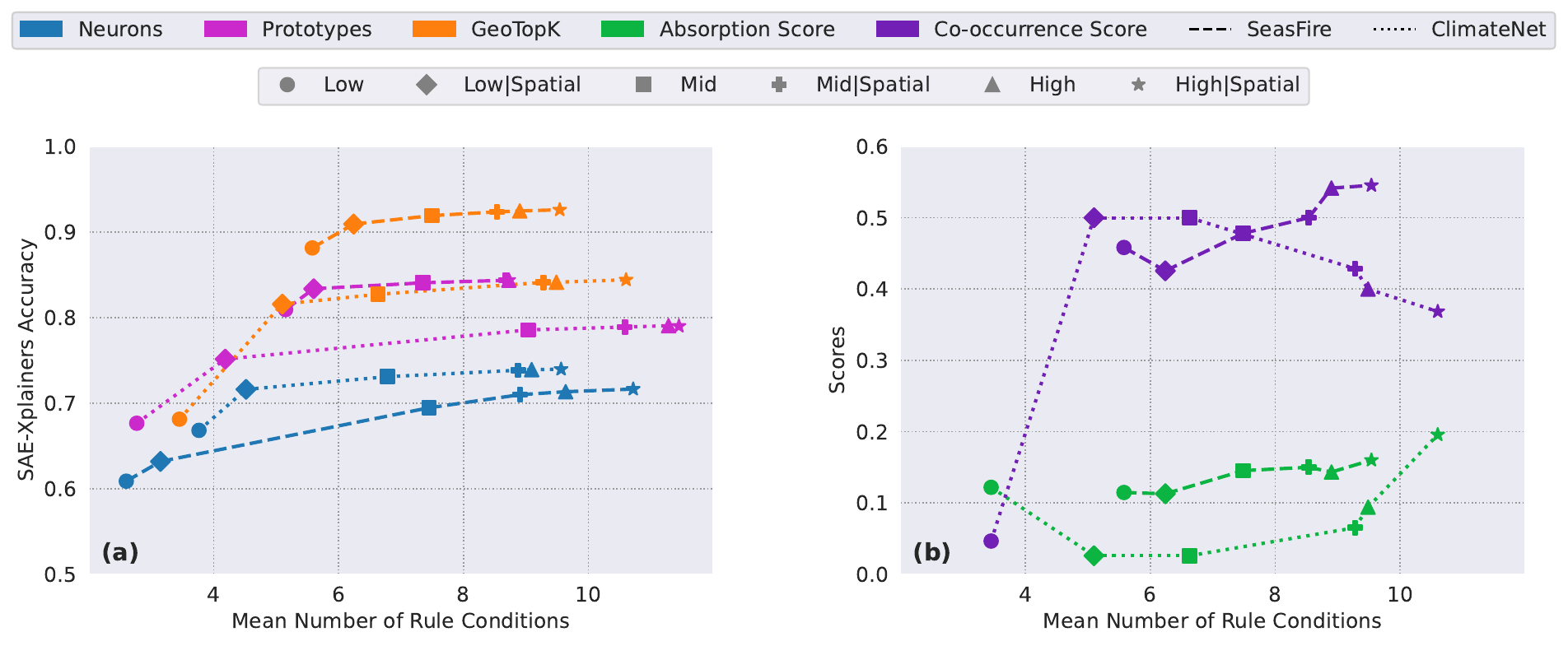}
    \caption{SAE-Xplainers quantitative analysis. \textbf{(a):} Faithfulness of our SAE-Xplainers interpretations across Skope-Rules parametrization of ClimaX Neurons, Prototypes, and GeoTopK feature activations. \textbf{(b):} Evolution of the scores related to absorption across SAE-Xplainers parametrizations.}
    \label{fig:xplain-quant}
\end{figure*}

\noindent\textbf{SAE-Xplainers Faithfulness.} 
In this section, we analyse two sets of SAE-Xplainers, learned over all the active features with enough positive samples of the SeasFire and ClimateNet datasets. For both datasets, we resample train/test splits to retain as many active features (in particular for ClimateNet). In both cases, we analyze the ClimaX model, as it outperforms ViT on both datasets. To assign a level of complexity to each feature and identify the importance of spatial context, we train our SAE-Xplainers for three parametrizations of Skope-Rules: low, mid, and high complexity, with or without spatial context. Then, for a specific feature we select the most suitable parametrization and spatial context that provides a gain in training accuracy of at least $\tau_{acc} = 0.01$ compared to the current setup, following this order of complexity [low, low and spatial, mid, ..., high and spatial]. This is reflected, in Figure~\ref{fig:xplain-quant}, by the six configurations per method and dataset showing an increase in the mean number of rule conditions across the SAE-Xplainers (details on SAE-Xplainers training, downstream model performance, and baselines are in the supplementary material).

In Figure~\ref{fig:xplain-quant}a, we compare the faithfulness via the prediction accuracy proxy of our SAE-Xplainers trained on the monosemantic SAE features to that of similar rule-based models trained on the polysemantic ClimaX model neurons and on class-specific prototypes, following the interpretable per-design prototype network: ProtoSeg \cite{sacha2023protoseg, porta2025multi}. We observe for the final aggregated sets of our SAE-Xplainers an evaluation accuracy of $\sim93\%$ ($+22\%$ with the neurons baseline and $+9\%$ with the prototypes) on SeasFire, and $\sim85\%$ ($+9\%$ with the neurons baseline and $+5\%$ with the prototypes) on ClimateNet. This shows first a strong faithfulness of the interpretations and validates our motivation to train accurate SAE-Xplainers on the GeoTopK features due to their monosemantic property as opposed to the model neurons. The faithfulness of our SAE-Xplainers, considering the adversarial sampling strategy used, also shows that GeoTopK does not only memorizes the geographic coordinates modulation, but uses it to disentangle the model representations. Additionally, we see that our SAE-Xplainers perform much better on SeasFire in comparison with ClimateNet, which can be explained by the higher dead features rate on ClimateNet (see Table~\ref{tab:main-result}), which causes superposition across the active SAE features, but also by the higher importance of spatial context in \textit{tropical cyclone} and \textit{atmospheric river} detection, contrarily to \textit{fire} forecasting. This is illustrated in Figure~\ref{fig:xplain-quant}a, with the large gain in model accuracy for our SAE-Xplainers on ClimateNet when integrating spatial context at low model complexity: $+13\%$, with only $+3\%$ for SeasFire. In supplementary material Figure 8, the feature assignment in SAE-Xplainers also demonstrates the role of spatial context in ClimateNet. 

\noindent\textbf{Feature Absorption.} We evaluate our SAE-Xplainers' ability to detect absorbed features for ClimaX fine-tuned on SeasFire and ClimateNet, using the same parametrizations as in Figure~\ref{fig:xplain-quant}a. As shown in Figure~\ref{fig:xplain-quant}b, on ClimateNet, excluding the low-complexity edge case where SAE-Xplainer accuracy is limited, the absorption score increases with rule accuracy, as it does for SeasFire. Moreover, for SeasFire, since \textit{fire} patterns are highly localized, more complex and accurate rules make co-occurrence of features with similar rule conditions more likely. For ClimateNet, however, ExEE patterns are far less localized, so co-occurrence is already high at low rule complexity, and the score decreases as complexity increases. The identification of feature absorption confirms the consistency between SAE-Xplainers' interpretations and SAE feature activations. Examples of absorbed features are provided in the supplementary material.

\begin{figure*}[ht]
  \centering
  \includegraphics[width=\linewidth]{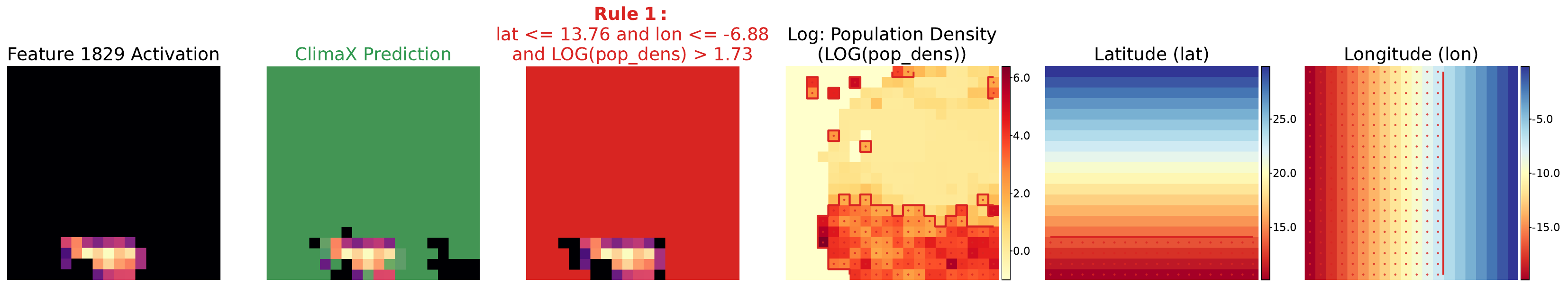}
   \caption{Sample-level feature interpretability for a ClimaX feature fine-tuned on SeasFire. \textbf{(Left $\rightarrow$ Right):} Feature activations, model prediction masks (\textit{no fires} regions are masked), SAE-Xplainer rule visualizations with binary masks (unsatisfied rule regions are masked), and input variables highlighting the rule support. The complete rule set for this feature is displayed in supplementary material Figure 11. 
   }
    \label{fig:sample}
\end{figure*}

\begin{figure*}[ht]
  \centering
  \includegraphics[width=\linewidth]{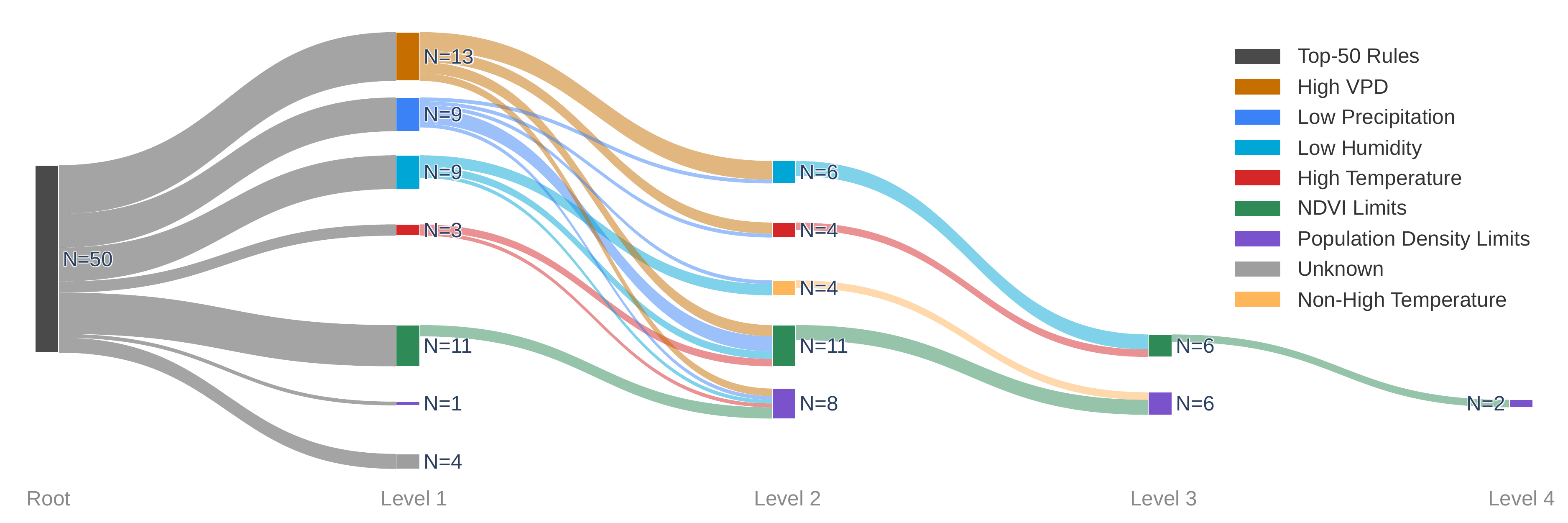}
   \caption{Semantic structure of the rules in the SAE-Xplainers of the 50 most fire-specific GeoTopK features. The levels represent the tree-like structure of each rule, while the nodes describe the categories of the conditions; \textbf{N} indicates the number of rules.}
    \label{fig:semantic}
\end{figure*}

\noindent\textbf{Samples Interpretation.} The ensemble of SAE-Xplainers enables local, sample-level interpretations, as illustrated in Figure~\ref{fig:sample}, where we analyze the activations of a ClimaX feature associated with \textit{fires} on a sample from the SeasFire dataset. The rule-based surrogate model of the feature first identifies the most relevant input variables, for which the corresponding rule boundaries can then be directly extracted. This provides structured 
insights to support human understanding of complex ExEE models at the sample level. Thanks to the high sparsity enforced during SAE training (with $k=4$), we limit each sample to at most four active features, ensuring that sample-level interpretations remain concise and tractable for human analysis. Our method also enables global model interpretability by projecting SAE features into a 2D space and describing them via the SAE-Xplainers, as illustrated by the exploratory tool\footnote{Available at https://eceo-epfl.github.io/SAE-Xplainers-umap} presented in the supplementary material. Additional sample analyses over SeasFire and ClimateNet are also included in the supplementary material and provide qualitative examples of scientifically consistent rule semantics.

\noindent\textbf{Scientific Semantics of SAE-Xplainers.} 
We assess whether the extracted rules encode scientifically meaningful concepts by analyzing, on SeasFire, the top-100 features of SAE-Xplainers on GeoTopK with the highest fire ratio for non-spatial rules across all complexity levels (ratio score and other selection algorithms detailed in the supplementary material). Table~\ref{tab:quant-cond} reports the proportion of rule sets containing established fire-risk conditions: high $\texttt{vpd}$, low $\texttt{tp}$, and low $\texttt{rel\_hum}$ \cite{kudlavckova2025assessing}, as \textbf{$\text{Tr}_{\text{cond}}$}, and their inverse (low-risk) as \textbf{$\text{Inv}_{\text{cond}}$}. SAE-Xplainers on GeoTopK contain literature-consistent conditions in 46\% of cases, while inverse conditions occur in only 2\%, compared with 16\% for ClimaX neurons and 20\% for prototypes, thus achieving semantic consistency comparable to the prototype interpretations while producing much fewer conflicting conditions.

\begin{table}[h!]
\centering
\begin{tabular}{c @{\hskip 8pt} c@{\hskip 6pt}c@{\hskip 6pt}c}
\toprule
\textbf{Metric}
& \textbf{Neurons} & \textbf{Prototypes} & \textbf{GeoTopK} \\
\cmidrule(lr){1-1} \cmidrule(lr){2-4}
$\text{Tr}_{\text{cond}} \uparrow$ & 0.33 & \textbf{0.50} & \underline{0.46} \\
$\text{Inv}_{\text{cond}} \downarrow$ & \underline{0.16} & 0.20 & \textbf{0.02} \\
\bottomrule
\end{tabular}
\caption{Ratio of rule sets containing true (Tr) and inverse (Inv) conditions for fire-specific dimensions.}
\label{tab:quant-cond}
\end{table}

We then examine the complete rule structure of the top-50 SAE-Xplainers. As shown in Figure~\ref{fig:semantic}, 92\% of the explanations contain conditions associated with at least one of seven literature-supported fire drivers (see supplementary material Table 8). For example, the \textbf{\textcolor{NDVIColor}{NDVI Limits}} filters out extreme NDVI values, as fires typically ignite and spread at mid-range NDVI \cite{krawchuk2009global}. The remaining 8\% are conservatively labeled as \textbf{\textcolor{UNKColor}{Unknown}} and reported in supplementary material Table 9. Overall, this targeted analysis shows that most fire-specific GeoTopK features admit compact interpretations grounded in established wildfire literature.  

\section{Conclusion} 

SAEs offer a promising explainability alternative for ExEE tasks, beyond standard methods focusing mainly on \textit{where} the model looks. We tackled two challenges related to using SAEs in ExEE research: the location-dependent patterns of the data, and the complex nature of input spaces for feature interpretation. We introduce GeoTopK, a location-aware SAE improving feature disentanglement, and SAE-Xplainers, a rule-based ensemble for feature interpretation. GeoTopK outperforms baselines across two ExEE datasets and transformer-based models, while SAE-Xplainers faithfully capture feature behaviors with semantic interpretations. Our approach could broadly generalize to other Earth system applications,  such as species distribution modeling or air quality forecasting. 
Future work can study the robustness of the extracted rules, which can be improved via constrained SAE training \cite{fel2025archetypal} or knowledge-grounded surrogate models, and explore richer surrogate architectures to better capture the spatiotemporal context.

\section{Acknowledgments}
We would like to thank Gaston Lenczner, IT engineer at ECEO, for developing the exploratory tool used in this work.

{\small
\bibliography{aaai2027}
}


\section{Implementation Details}
\label{sec:detail}
The code for this paper is available at \url{https://github.com/eceo-epfl/SAE-Xplainers}. The experiments are all done on an NVIDIA GeForce RTX 3090 Ti GPU with 24 GB of memory, and an Intel Xeon W-2275 CPU with 125 GB of memory on Ubuntu 22.04. If not specified, all runs are done once, and as shown in the code all seeds are set using PyTorch Lightning \texttt{seed\_everything} function.
\subsection{Dataset Descriptions}
\label{sec:data-desc}

SeasFire is a global dataset for dense wildfire forecasting across multiple horizons (from 8 days to 128 days, in 8-day intervals). In this paper, we focus on the shorter 8-day horizon to ensure strong performance of the analyzed models: ClimaX and ViT, without the need to integrate the temporal dimension (leading to a more complex interpretation of the models' behavior). SeasFire spans 21 years from 2001 to 2021 with aggregated input predictors over 8-day windows and a spatial grid of resolution $0.25\degree$ from multiple sources, as detailed in Table~\ref{tab:predictors}. Samples in SeasFire are constructed as $80 \times 80$ images extracted from the global Earth representation, where at least one fire occurs, leading to 44k samples over 17.7M patches ($4 \times 4$ pixels), with 14 input variables as described in Table~\ref{tab:predictors}. $98\%$ of those patches are assigned to the \textit{no fire} class, causing the strong imbalance.

ClimateNet is a global dataset for the detection of tropical cyclones and atmospheric rivers via human-expert annotations on daily simulations of a climate model: CAM5.1, from 1996 to 2013. In ClimateNet, samples represent Earth as a whole in images of spatial size $768 \times 1152$ pixels with $6$ input predictors (see Table~\ref{tab:predictors}), at a spatial resolution of $25$ km. This constitutes an important difference w.r.t. SeasFire, as the detection tasks leverage a large spatial context compared to wildfire forecasting. Therefore, ClimateNet complements SeasFire in two ways: it focuses on two different events, each one triggered by specific patterns and requiring the capture of long-range phenomena. ClimateNet is composed of 459 samples over 25.4M patches, with $94\%$ of \textit{no event} patches.

Our selection of these datasets is motivated first by the growing danger and frequency of these ExEEs under climate change, and second by their data characteristics: although they share input-space similarities that raise similar challenges for SAE feature interpretation, they differ substantially in task formulation, sample construction, and W\&C data generation (reanalysis vs. simulation).

\begin{table*}[t]
\centering
\setlength{\tabcolsep}{1mm}
\begin{tabular}{c @{\hskip 8pt} c@{\hskip 6pt}c@{\hskip 6pt}c@{\hskip 6pt}c@{\hskip 6pt}c}
\toprule
\textbf{Dataset} & \textbf{Acronym} & \textbf{Name} & \textbf{Units} &\textbf{Type} & \textbf{Source} \\
\cmidrule(lr){1-1} \cmidrule(lr){2-6}
\multirow{14}{*}{SeasFire}
& \texttt{lst\_day}  & Land Surface Temperature (Day) & K     & Land Surface (Weather)  & MODIS \\
& \texttt{ndvi}      & Normalized Difference Vegetation Index & unitless & Vegetation  & MODIS \\
& \texttt{mslp}      & Mean Sea Level Pressure      &   Pa   & Weather     & ERA5  \\
& \texttt{rel\_hum}  & Relative Humidity            &   $\%$      & Weather     & ERA5  \\
& \texttt{ssrd}      & Surface Solar Radiation Downwards & $\text{MJ}\cdot\text{m}^{-2}$  & Weather     & ERA5  \\
& \texttt{sst}       & Sea Surface Temperature &     K        & Climate     & ERA5  \\
& \texttt{swvl1}     & Volumetric Soil Water Layer 1 & unitless       & Land Surface (Weather)        & ERA5  \\
& \texttt{t2m\_mean} & 2m Temperature Mean      & K             & Weather     & ERA5  \\
& \texttt{tp}        & Total Precipitation      & m            & Weather     & ERA5  \\
& \texttt{vpd}       & Vapor Pressure Deficit   & hPa            & Weather     & ERA5  \\
& \texttt{ws10}      & Wind Speed at 10m        & $\text{m} \cdot \text{s}^{-1}$            & Weather     & ERA5  \\
& \texttt{pop\_dens} & Population Density       & $\text{persons} \cdot\text{km}^{-2}$            & Anthropogenic & GPWv4 \\
& \texttt{lat}       & Latitude                 & degrees            & Geographic  & -     \\
& \texttt{lon}       & Longitude                & degrees            & Geographic  & -     \\
\cmidrule(lr){1-1} \cmidrule(lr){2-5}
\multirow{6}{*}{ClimateNet}
& \texttt{TMQ}  & Total Vertically Integrated Precipitable Water &  $\text{kg} \cdot \text{m}^{-2}$ & Weather    & CAM5.1 \\
& \texttt{U850} & Zonal Wind at 850 mbar       &    $\text{m} \cdot \text{s}^{-1}$              & Weather    & CAM5.1 \\
& \texttt{V850} & Meridional Wind at 850 mbar   &   $\text{m} \cdot \text{s}^{-1}$              & Weather    & CAM5.1 \\
& \texttt{PSL}  & Sea Level Pressure         & Pa                   & Weather    & CAM5.1 \\
& \texttt{lat}  & Latitude                   & degrees                   & Geographic & -      \\
& \texttt{lon}  & Longitude                  & degrees                   & Geographic & -      \\
\bottomrule
\end{tabular}
\caption{Summary of predictors used from SeasFire and ClimateNet datasets in our paper.}
\label{tab:predictors}
\end{table*}

\subsection{Models fine-tuning}
\label{sec:model-fin}

For both datasets, ClimaX is independently fine-tuned, leveraging its pretrained weights at $\sim 1.4 \degree$. On top of the last layer of ClimaX ($h_t$), we train a lightweight MLP decoder with two hidden layers. The set of pre-training variables from ClimaX only partially overlapped with the predictors from SeasFire and ClimateNet. As a consequence, for the initialization of the variable-specific embedding and projection, when there is no direct match in the pre-training set for SeasFire and ClimateNet predictors, we manually matched them to the closest (or the group of closest) pre-training variables. For instance, the \texttt{vpd} projection and embedding in SeasFire is initialized with the mean of the weights from the following pre-training variables: relative humidity and temperature at 2 meters. Finally, if no group of pre-training variables would match, we initialized the variable-specific weights with the average across all pre-training variables. Experimentally, we observe that, when trained from scratch, ClimaX only reaches $\sim 0.5$ PRAUC on SeasFire compared to the results in Table~\ref{tab:downstream}, highlighting the benefits of fine-tuning from its pre-trained weights.

On both datasets, ClimaX is fine-tuned via the ADAMW optimizer and a cosine annealing scheduler with linear warm-up. The learning rate is also adjusted for the ClimaX backbone compared to the first projection heads and the decoder by a factor of $0.2$. For the hyperparameters, we directly reuse the ones leveraged in the ClimaX fine-tuning use case from \cite{nguyen2023climax} with learning rate: $5e-4$, warm-up learning rate: $1e-8$, betas: $[0.9, 0.999]$, and weight decay: $1e-5$. We fine-tune across a total of $50$ epochs with $5$ for warm-up and a batch size of $64$ for SeasFire and $8$ for ClimateNet (due to the large image size for this dataset). Across both datasets, fine-tuning with an adjusted learning rate always outperforms freezing the backbone weights. We use standard cross-entropy loss for SeasFire and cross-entropy combined with dice loss on ClimateNet, after testing cross-entropy and dice loss alone. During fine-tuning on ClimateNet, we randomly center crop to a size of $768 \times 768$ due to computational constraints, but run inference on the original size of $768 \times 1152$. The selection of the fine-tuning hyperparameters is done following the criterion presented in Table~\ref{tab:downstream}. Nonetheless, the objective of our paper is not to improve downstream performance on either dataset, but rather to remain close to the original papers' reported baselines while focusing on the interpretability of the resulting models.

A ViT backbone is also trained from scratch on both datasets with the same lightweight MLP decoder as ClimaX. We leverage similar training hyperparameters as with ClimaX for the optimizer and learning rate scheduler, without adjusting the learning rate for the backbone. Experimentally, we just adapt the learning rate to $1e-5$ for SeasFire and $5e-5$ for ClimateNet by testing across the following values: $[5e-4, 1e-4, 5e-5, 1e-5]$, as we observe overfitting at $5e-4$. The loss is the same for SeasFire, but on ClimateNet we experimentally found that weighted cross-entropy combined with dice outperforms standard cross-entropy. The weights selected are $0.2$ for the background class and $1$ for both event classes. The results in Table~\ref{tab:downstream} show that ClimaX outperforms the ViT on both datasets. Therefore, while SAE Performance Analysis is conducted on both models across SeasFire and ClimateNet, for the Rule-based Feature Interpretations of our SAE-Xplainers we focus on ClimaX only.

\begin{table*}[t]
\centering
\begin{tabular}{c @{\hskip 8pt} c@{\hskip 6pt}c@{\hskip 6pt}c@{\hskip 6pt}c @{\hskip 8pt} c@{\hskip 6pt}c@{\hskip 6pt}c@{\hskip 6pt}c}
\toprule
\multirow{2}{*}{\textbf{Metric}}
& \multicolumn{4}{c}{\textbf{SeasFire}}
& \multicolumn{4}{c}{\textbf{ClimateNet}} \\
\cmidrule(lr){2-5} \cmidrule(lr){6-9}
& \textbf{ClimaX} & \textbf{ViT} & \textbf{ProtoSeg} & \textbf{U-Net++*}
& \textbf{ClimaX} & \textbf{ViT} & \textbf{ProtoSeg} & \textbf{DeepLabV3+*} \\
\cmidrule(lr){1-1} \cmidrule(lr){2-5} \cmidrule(lr){6-9}
PRAUC $\uparrow$ & \textbf{0.6205} & 0.6150 & 0.5161 & 0.6100 & - & - & - & - \\
IoU $\uparrow$   & - & - & - & - & \textbf{0.5385} & 0.4972 & 0.4254 & 0.5247 \\
\bottomrule
\end{tabular}
\caption{Downstream task performance of ClimaX, ViT, and ProtoSeg fine-tuned on SeasFire and ClimateNet. * denotes results reported from the original papers.}
\label{tab:downstream}
\end{table*}

\subsection{SAE Training}
\label{sec:sae-train}
On both datasets for ClimaX and ViT, we train a single linear-layer SAE with the TopK activation and a batch size of $256$. To guarantee that the TopKSAE activations are non-negative, we add a ReLU function in the encoder as in \cite{gao2024scaling}. The geolocation encoder in GeoTopK is composed of an MLP with two hidden layers. All our SAEs are trained via mean squared error and an auxiliary loss term to decrease the dead features rate. We train for $20$ epochs on SeasFire, and $60$ epochs on ClimateNet, as the $\text{R}^2$ and dead features rate criterion converge at different speeds across the datasets. The auxiliary loss term is an adaptation of ghost gradient for TopK SAE from the Overcomplete library. The pseudo-code for the auxiliary loss term is available in Algorithm~\ref{alg:auxk}. Prior to computing the loss in Equation 2, the reconstructed patches $\hat{h}_{t, \text{mod}}$ are projected back to the original activation space via the following inversion formula: 

\begin{equation}
    \hat{h}_t = \frac{\hat{h}_{t, \text{mod}} - \tanh(\beta)}{1 + \tanh(\alpha) + \epsilon} \; ,
\end{equation}
with $\epsilon = 1e-8$. We leverage the ADAM optimizer with no scheduler to train the SAEs. For ClimaX fine-tuned on SeasFire, we test for TopK and GeoTopK, multiple levels of regularization for the auxiliary loss $\gamma \in [0.1, 0.01, 0.001]$ at a fixed learning rate: $5e-4$, and identify $\gamma = 0.01$ as the best for both SAEs. For ClimaX fine-tuned on ClimateNet, we apply a similar strategy while also adapting the learning rate due to overfitting, finding as the best regularization: $\gamma = 0.1$ for TopK, with a learning rate: $5e-5$, and for GeoTopK: $\gamma = 0.01$ and a learning rate: $5e-6$. The hyperparameters selection is based on the $\text{R}^2$ and dead features rate criterion; when $\text{R}^2$ reaches a high value, we focus on optimizing the dead features rate, as SAEs can often reach a degenerate optimum with high reconstruction performance but only a handful of active features.

Moreover, to decrease the dead features rate we tie at initialization the encoder and decoder weights: $W_{enc} = W_{dec}^{T}$ following \cite{huben2024sparse} and, as the rate remains high on ClimateNet, we employ the feature resampling strategy from \cite{bricken2023monosemanticity} to further decrease it by resampling every epoch across 4.5 M patches.

For the ViT models on both datasets, we directly reuse the best hyperparameters selected for ClimaX. The impact of the latent dimension size on the SAEs' performance is detailed in subsection Scaling SAE Latent Dimension.


\begin{algorithm}[h]
\caption{Top-$K$ Auxiliary Loss (AuxK)}
\label{alg:auxk}
\begin{algorithmic}[1]

\Require Input $h_t$, reconstruction $\hat{h}_t$, pre-activations $z_{\text{pre}}$, activations $z$, dictionary $W_{dec}$

\State $r \leftarrow h_t - \hat{h}_t$ \Comment{Compute residual}

\State $z_{\text{pre}} \leftarrow \text{ReLU}(z_{\text{pre}}) - z$ \Comment{Isolate unchosen feature activations}

\State $\mathcal{I} \leftarrow \text{TopK}\!\left(z_{\text{pre}},\ k = \lfloor N/2 \rfloor\right)$ \Comment{Select top-$50\%$ unchosen features}
\State $\tilde{z} \leftarrow 0;\quad \tilde{z}_{\mathcal{I}} \leftarrow z_{\text{pre},\,\mathcal{I}}$ \Comment{Build sparse auxiliary vector}

\State $\hat{r} \leftarrow W_{dec}\,\tilde{z}$ \Comment{Predict residual from auxiliary features}

\State $\mathcal{L}_{\text{aux}} \leftarrow \|r - \hat{r}\|^2_2$

\Return $\mathcal{L}_{\text{aux}}$
\end{algorithmic}
\end{algorithm}

\subsection{SAE-Xplainers Training} 
\label{sec:saeX-train}

The attention scores leveraged in Equation 8 are aggregated by averaging across all layers up to $l$: $W_{\text{att}}^{l} = \frac{1}{B} \sum_{i=1}^BA_{i}$ with $B$ the number of attention blocks and $A_i \in \mathbb{R}^{T \times T}$ the attention matrix. Then patch-level scores are $A^{l}(x_t) = [W_{\text{att}}^{l}]_{t,:}$.

We train our set of SAE-Xplainers on features with enough active samples for both datasets. On SeasFire, due to a low dead features rate and the overall high activation counts of the learned features, we target dropping below $1\%$ of them by applying a threshold of 1600 active samples across train and evaluation sets (test and validation). On the contrary, with ClimateNet, given our high dead features rate and low activation counts, we target dropping around $10\%$ of the features with a threshold of 160 to ensure sufficient samples for training the surrogate models; as a comparison in \cite{huben2024sparse}, the LLM autointerpretability framework only uses 20 samples of 64-token sentence-fragments. Following the top-random balanced sampling strategy of \cite{huben2024sparse, paulo2025automatically}, active patches are sampled by selecting the most activated half, and the remaining half are randomly sampled from the active distribution. For the neuron and prototype baselines, we first binarize patch activations by thresholding at the $99^{th}$ percentile of each dimension, then apply the same top-random balanced strategy.

In Table~\ref{tab:skope_rules}, we detail the Skope-Rules parametrization for the different complexity levels explored for the training of the SAE-Xplainers. Skope-Rules is a Python machine learning module that extracts interpretable \texttt{if–then} logical rules from an ensemble of decision trees, weighting each rule by its out-of-bag precision. For both datasets, as they are global, we trained with a radius $\sigma = 1000$ km as a good range to find proper negatives still close in local environmental patterns. We evaluate the adversarial sampling strategy against standard random sampling in the subsection Sampling Analysis. We leverage for both datasets a spatial context of $20$ patches. 

\begin{table}[h]
\small
\centering
\setlength{\tabcolsep}{1mm}
\begin{tabular}{c @{\hskip 8pt} c@{\hskip 6pt}c@{\hskip 6pt}c@{\hskip 6pt}c}
\toprule
\multirow{2}{*}{\textbf{Levels}}
& \multicolumn{4}{c}{\textbf{Parameters}} \\
\cmidrule(lr){2-5}
& \textbf{Max Depth} & \textbf{\# Estim.} & \textbf{Min Precision} & \textbf{Min Recall} \\
\cmidrule(lr){1-1} \cmidrule(lr){2-5}
Low  & 3 & 3 & 0.90 & 0.20 \\
Mid  & 4 & 5 & 0.80 & 0.15 \\
High & 5 & 5 & 0.70 & 0.10 \\
\bottomrule
\end{tabular}
\caption{Skope-Rules parametrization across complexity levels. \#~Estim. denotes the number of estimators.}
\label{tab:skope_rules}
\end{table}

\subsection{Absorption Scores}
For the co-occurrence score, we select the $M=50$ closest pairs and set the co-occurrence threshold to $\tau = 0.1$. Additionally, we consider two rule-condition threshold values equivalent if their difference is within $5 \%$ of the target variable's standard deviation.

For the SAEBench absorption score, we reuse the same probe and $\ell_1$-probe hyperparameters as in \cite{karvonen2025saebench}, and set the cosine similarity threshold to $0.0$ and the F1 threshold to $0.2$.

\begin{table*}[t]
\centering
\begin{tabular}{c @{\hskip 8pt} c@{\hskip 6pt}c@{\hskip 6pt}c@{\hskip 6pt}c@{\hskip 6pt}c@{\hskip 6pt}c}
\toprule
\textbf{Metric}
& \textbf{Scalar} & \textbf{Shift-Only} & \textbf{Scale-Only} & \textbf{LOAN} & \textbf{LatBias} &\textbf{GeoTopK}\\
\cmidrule(lr){1-1} \cmidrule(lr){2-7}
${\text{R}^2}$ $\uparrow$              & 0.930 & \underline{0.936} & 0.931 & 0.754 & 0.928 & \underline{0.937} \\
$\text{R}^2_{\text{worst}}$ $\uparrow$ & 0.884 & \underline{0.899} & 0.889 & 0.616 & 0.881 & \underline{0.901} \\
$\text{R}^2_\text{event}$ $\uparrow$   & 0.808 & 0.802 & \textbf{0.817} & 0.726 & 0.806 & 0.807 \\
MSE $\downarrow$                        & 1.77e-3 & \underline{1.61e-3} & 1.73e-3 & 0.243 & 1.82e-3 & \underline{1.59e-3}\\
Dead Features $\downarrow$              & 0.041 & \underline{0.004} & \underline{0.003} & \underline{0.002} & 0.107 & \underline{0.002}\\
\bottomrule
\end{tabular}
\caption{Ablation study of the geographic modulation in GeoTopK on ClimaX fine-tuned on SeasFire. \textbf{Bold} means best score, \underline{underline} means similar score (difference is $\leq 0.005$).}
\label{tab:ablation}
\end{table*}

\begin{figure*}[t]
    \centering
    \includegraphics[width=0.6\textwidth]{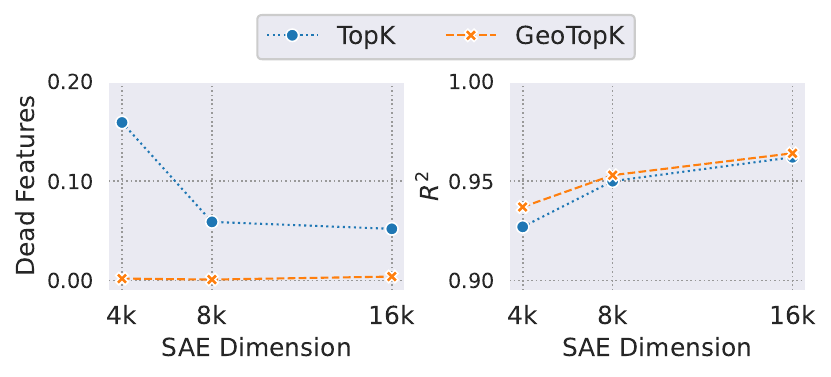}
    \caption{TopK and GeoTopK performance (\textbf{Left:} Dead Features Rate, \textbf{Right:} ${\text{R}^2}$) when scaling the latent dimension size: $N$, at fixed sparsity ratios for ClimaX, fine-tuned on SeasFire.}
    \label{fig:scaling}
\end{figure*}

\subsection{Prototype Network Training}
\label{sec:proto-train}

To study how our Skope-Rules sets compare against a different family of interpretable concept-extraction methods, we train a dense prediction prototype network, ProtoSeg \cite{sacha2023protoseg, porta2025multi}, on top of ClimaX fine-tuned on SeasFire and ClimateNet. Since ClimaX is already fine-tuned on the target task, we skip ProtoSeg's joint training stage and run only the warm-up and fine-tuning stages: the prototype layer is trained during warm-up with the decoder head frozen, and the decoder head is subsequently trained once the prototypes have been pushed to real patch samples.

We reuse the ClimaX fine-tuning hyperparameters, with two exceptions: we replace AdamW with Adam and omit the learning-rate scheduler, since we train for substantially fewer epochs (10 per stage). We also do not apply the adjusted learning rate, as the backbone is frozen throughout. For the loss, we use cross-entropy for SeasFire and a combined cross-entropy/dice loss for ClimateNet, consistent with the best-performing configurations reported for ClimaX. For regularization, we adopt the losses defined in ProtoSeg: a Kullback–Leibler divergence term weighted at $0.25$ and an $\ell_1$ penalty on the decoder head weighted at $1e-4$.

We evaluate two distance formulations for the prototype layer: the standard Euclidean distance paired with ProtoSeg's logarithmic activation, and a cosine distance paired with a linear activation. On SeasFire, the cosine-distance variant outperforms the standard formulation, so we adopt it for all subsequent experiments. Finally, we use 20 prototypes per class, yielding 40 prototypes total for SeasFire (2 classes) and 60 for ClimateNet (3 classes).

\section{GeoTopK Additional Results}
\subsection{Ablation Study}
\label{sec:abl}

In our method, GeoTopK is parametrized via a feature-wise affine modulation. Through the ablation study in Table~\ref{tab:ablation}, we compare the performance of our GeoTopK with other potential parametrizations of the geographic location modulation. We train on ClimaX fine-tuned for SeasFire a scalar modulation with $(\alpha, \beta) \in \mathbb{R}^{2}$, a shift-only modulation with $\alpha = 1$ and $\beta \in \mathbb{R}^{D}$, and a scale-only modulation with $\alpha \in \mathbb{R}^{D}$ and $\beta = 0$. We additionally compare against location-aware adaptive normalization (LOAN) \cite{eddin2023location} and per-latitude SAE band bias (LatBias). The results show that, across all metrics, GeoTopK parametrization performs best except on $\text{R}^2_{\text{event}}$, and its performance is close to that of the shift-only parametrization. Computing the MSE loss in the modulated space leads to a trivial solution with zero scale and shift vectors ($\text{R}^2 < 0$ on SeasFire). Finally, $\tanh$ bounds the scale and shift magnitudes, unlike GELU or ReLU, to avoid that the SAE only learns to represent geographic locations.

\subsection{Scaling SAE Latent Dimension}
\label{sec:scaling}

We study whether the benefits of GeoTopK persist as we increase the latent dimension size of the SAE. Focusing on ClimaX for SeasFire, we scale this size from $N: 4096 \rightarrow 16384$ at a fixed sparsity ratio $k: 4 \rightarrow 16$ (Figure~\ref{fig:scaling}). We observe on the SeasFire test that with more capacity (higher $k$ and $N$), the TopK baseline performance converges towards GeoTopK with similar ${\text{R}^2}$. However, the baseline consistently suffers from a significantly higher percentage of dead features. We hypothesize that larger TopK models can explicitly approximate the geographic priors encoded in GeoTopK, but do so less efficiently. GeoTopK achieves better results with fewer parameters, enabling high-performance SAEs at a reduced computational cost.

\begin{figure*}[t]
  \centering
  \begin{subfigure}{\columnwidth}
    \includegraphics[width=\linewidth]{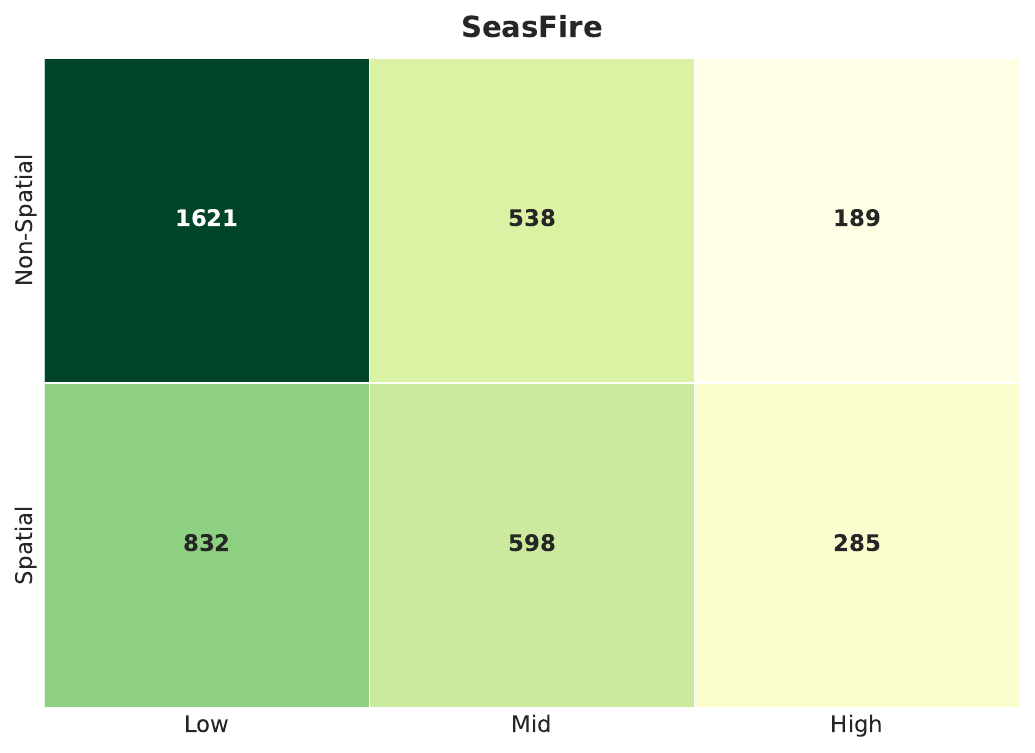}
  \end{subfigure}
  \hfill
  \begin{subfigure}{\columnwidth}
    \includegraphics[width=\linewidth]{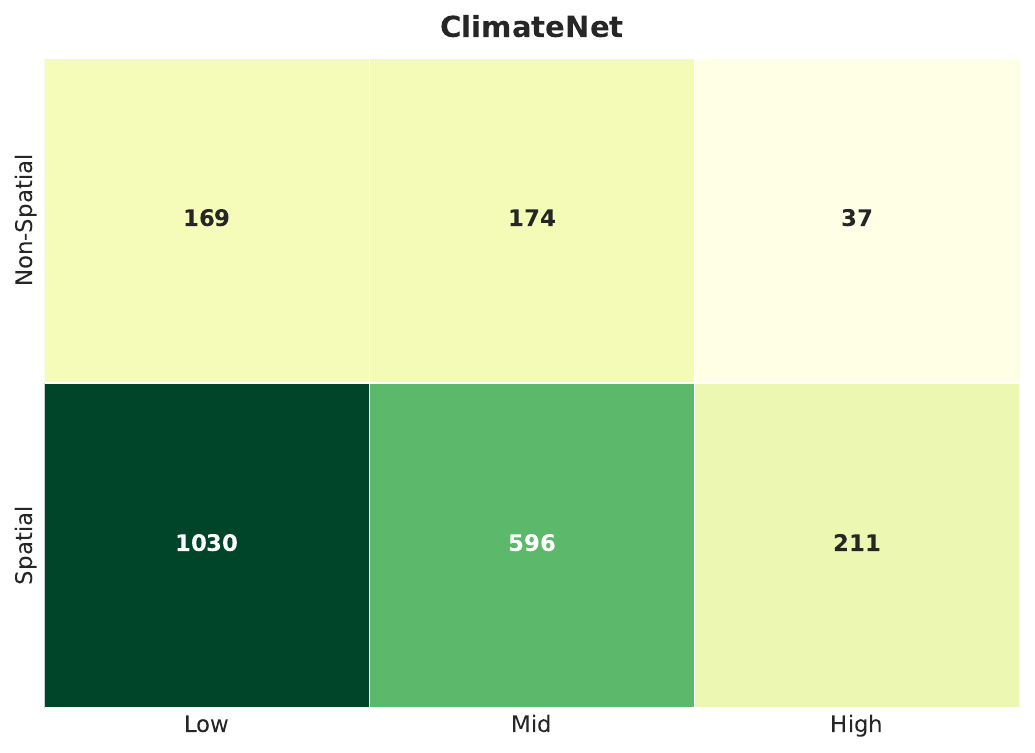}
  \end{subfigure}
  \caption{Feature assignment across the Skope-Rules parametrization: low, mid, and high, with or without spatial context.}
  \label{fig:assign}
\end{figure*}

\begin{figure*}[h!]
  \centering
  \includegraphics[width=\linewidth]{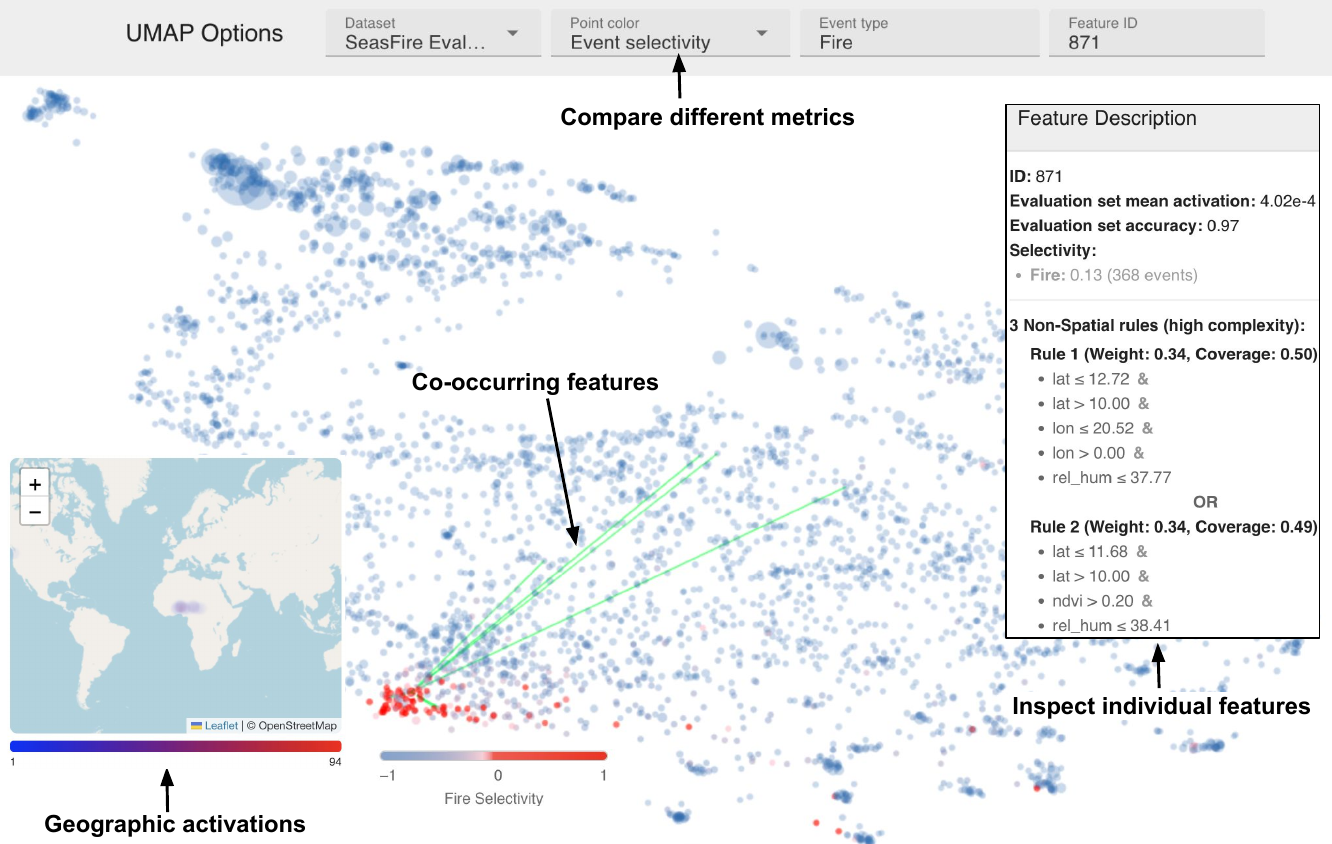}
  \caption{Exploratory tool for global model interpretations showcasing feature vectors as points in 2 dimensions extracted through a UMAP projection, and color-coded by event class selectivity. When a feature is selected, its rule-based interpretations, geographic activations, and co-occurring features are shown.}
  \label{fig:demo}
\end{figure*}

\section{SAE-Xplainers Additional Results}
\subsection{Features Assignment}
\label{sec:assign}

We show in Figure~\ref{fig:assign} the different patterns between SeasFire and ClimateNet in feature assignment for the SAE-Xplainers. On SeasFire, the spatial dimension has less importance than for ClimateNet, where a majority of the features are assigned to Skope-Rules parametrization with spatial context. This can be explained by the types of extreme events and tasks tackled in ClimateNet, which are inherently more spatial, but also by the difference in the number of predictors available in both datasets: 14 vs 6, as such, each patch in SeasFire contains more information.

\subsection{Sampling Analysis}
\label{sec:sampl}

The sampling for the training of the SAE-Xplainers directly impacts their observed faithfulness and the feature interpretation quality. We analyze, across three runs of SAE-Xplainers for ClimaX fine-tuned on SeasFire, three metrics quantifying feature interpretation quality. We focus on the most used parametrization setup in the final ensemble of SAE-Xplainers: low, without spatial context (see Figure~\ref {fig:assign}). We first assess if the interpretability of features remains consistent across independent rule sets by calculating the \textbf{Stability}: it measures, for each feature, the pairwise average Jaccard distance between the rule sets across runs, using the Hungarian algorithm to match corresponding rules. The \textbf{$\text{Weight}_{\text{loc}}$} allows us to study the importance of the geographic coordinates in feature interpretations by measuring the ratio of rules with a condition on either latitude or longitude. Finally, the \textbf{$\text{Tr}_{\text{cond}}$} assesses the agreement between model's prediction and rule interpretation by measuring the proportion of the top-100 fire-specific features interpreted by rules that contain known patterns leading to high fire risk: high $\texttt{vpd}$, low $\texttt{tp}$ and low $\texttt{rel\_hum}$, as defined in the Section Rule-based Feature Interpretations.

Results in Table~\ref{tab:xplainers} shows that both sampling strategies lead to similar stability, with an improvement in rules quality for the adversarial sampling: the rules focus on fire drivers other than the geographic coordinates (lower \textbf{$\text{Weight}_{\text{loc}}$}) and both their interpretation are in agreement with known and interpretable fire patterns (high \textbf{$\text{Tr}_{\text{cond}}$}).

\begin{table}[h]
\centering
\small
\begin{tabular}{c @{\hskip 8pt} c@{\hskip 6pt}c@{\hskip 6pt}c}
\toprule
\multirow{2}{*}{\textbf{Sampling}}
& \multicolumn{3}{c}{\textbf{Metric}} \\
\cmidrule(lr){2-4}
& \textbf{Stability $\downarrow$} & \textbf{$\text{Weight}_{\text{loc}}$ $\downarrow$} & \textbf{$\text{Tr}_{\text{cond}} \uparrow$} \\
\cmidrule(lr){1-1} \cmidrule(lr){2-4}
Random   & $\textbf{0.688} \pm 0.023$ & $0.633 \pm 0.059$ & $0.490 \pm 0.064$ \\
Adversarial & $0.705 \pm 0.017$ & $\textbf{0.574} \pm 0.071$ & $\textbf{0.497} \pm 0.074$ \\
\bottomrule
\end{tabular}
\caption{Comparison of the sampling strategies for training the SAE-Xplainers on ClimaX fine-tuned on SeasFire. The reported results are average across three runs with their standard deviation.}
\label{tab:xplainers}
\end{table}

\subsection{Global Model Interpretations.}
\label{sec:glob}

GeoTopK and SAE-Xplainers can be leveraged in synergy for global model analysis on ExEE tasks. To support this, we developed an exploratory tool inspired by \cite{bohacek2025uncovering} (see Figure~\ref{fig:demo}). The tool enables different visualizations that domain experts can build upon to derive insights: a UMAP projection of SAE dictionary vectors $d_i$ into two dimensions to visualize feature similarity and enable cluster analysis of related concepts, interactive access to feature co-occurrences, feature descriptions with detailed rule information from SAE-Xplainers, and geographic activations. 
Leveraging the interpretability enabled by GeoTopK and SAE-Xplainers, the tool provides ExEE experts with intuitive visualizations that can be turned into climatic insights. The exploratory tool is available online at \url{https://eceo-epfl.github.io/SAE-Xplainers-umap} and focuses on \textbf{non-spatial rules} only.

\subsection{Sample Interpretations}
\label{sec:plot-samp}

In Figure~\ref{fig:samp-fire-1} and~\ref{fig:samp-fire-2}, we provide other instances of sample-level interpretations for features linked to \textit{fires} for ClimaX fine-tuned on SeasFire. Moreover, in Figure~\ref{fig:samp-ar-1} and~\ref{fig:samp-ar-2} we introduce examples of sample-level interpretations for features linked to \textit{atmospheric rivers} for ClimaX fine-tuned on ClimateNet, where we zoom in on a specific window of high feature activations for the given sample. 
These additional visualizations show the high interpretability of our SAE-Xplainers over different datasets and features.

\subsection{Absorbed Features}
\label{sec:plot-abs}

In Figure~\ref{fig:group} and~\ref{fig:group-2}, we share examples of groups of absorbed features on the ClimateNet and SeasFire datasets. In Figure~\ref{fig:group}, Feature $3696$ is the main feature, partially absorbed when it co-occurs with Feature $1670$. The SAE-Xplainers' interpretations are very similar for those features, while they partially co-occurred. This caused their selection to compute the SAEBench absorption score, defining the prediction task with the rule-based surrogate model assigned to the main feature.

\subsection{Details on SAE-Xplainers Semantic Consistency.}
\label{sec:det-sem}

In this section, we provide details and extend the analysis of the SAE-Xplainers semantic meaning for the fire forecasting use-case. First, we introduce the selection score for each dimension type: neurons, prototypes, and GeoTopK features, used in Table 2 and Figure 6. For prototypes, we simply select the ones assigned in the first training stage to the class \textit{fire} before the push stage, as described in \cite{sacha2023protoseg, porta2025multi}. The neurons and SAE features are selected as dimensions with the highest \textit{fire} class ratio $\mathcal{R}_{\textit{fire}, i}$. The feature ratio for the target class \textit{fire} is computed over the SeasFire test set with $z_i(x_t)$ the activation of feature $d_i$ for the input patch sample $x_t$, and $\mathcal{C^{*}} = \{x_t \mid y_t=c^{*}, \: x \in \mathcal{X}\}$ the set of input patch samples $\mathcal{X}$ assigned to the target class $c^{*}$: 

\begin{equation}
    \mathcal{R}_{c^{*}, i} = \frac{\sum_{x_t \in \mathcal{C^{*}}} |z_i(x_t)|}{\sum_{x_t \in \mathcal{X}} |z_i(x_t)| + \epsilon}\;.
\end{equation}

For GeoTopK, we have $|z_i(x_t)| = z_i(x_t)$ due to the TopKSAE encoder that internally contains a ReLU layer.

\begin{table*}[tp]
\centering
\setlength{\tabcolsep}{6pt}
\begin{tabular}{C{2.5cm} C{1.cm} C{2cm} L{5.5cm} C{4.5cm}}
\hline

\textbf{Category} & \textbf{Usage} & \textbf{Ex. Feature} & \textbf{Description} & \textbf{Reference} \\ \hline

\textbf{\textcolor{NDVIColor}{NDVI Limits}} & 56\% & 632 &
NDVI, a proxy for vegetation productivity, shows a humped relationship with fire risk: both sparse and overly healthy (green) vegetation suppress it.
& \cite{krawchuk2009global, hantson2016status, bowman2020vegetation} \\ \hline

\textbf{\textcolor{POPColor}{Population Density Limits}} & 34\% & 3734 &
Fire risk peaks at intermediate population density, the wildland–urban interface, since fire ignition rises but fire spread falls with density, producing a humped relationship.
& \cite{calef2008human, hantson2016status, pechony2009fire, li2012process} \\ \hline

\textbf{\textcolor{LowHumidity}{Low Humidity}} & 30\% & 49 &
Low relative humidity dries fuel and soil, favoring ignition and spread, and is thus negatively correlated with fire risk.
& \cite{li2012process, jolly2015climate, jain2022observed, kudlavckova2025assessing} \\ \hline

\textbf{\textcolor{HighVPD}{High VPD}} & 26\% & 3815 &
Vapor pressure deficit (VPD): air dryness at a given temperature, drives fuel aridity and is positively correlated with ignition and spread.
& \cite{pechony2009fire, jain2022observed, kudlavckova2025assessing} \\ \hline

\textbf{\textcolor{LowPrecip}{Low Precipitation}} & 18\% & 2728 &
Precipitation, closely tied to humidity, sustains fuel and soil moisture, and is thus negatively correlated with fire risk.
& \cite{hantson2016status, jolly2015climate, kudlavckova2025assessing} \\ \hline

\textbf{\textcolor{HighTemp}{High Temperature}} & 14\% & 700 &
High temperature promotes fuel aridity and is generally positively correlated with fire risk.
& \cite{hantson2016status, bowman2020vegetation, jolly2015climate, jain2022observed} \\ \hline

\textbf{\textcolor{NonHighTemp}{Non-High Temperature}} & 8\% & 1910 &
Extreme temperature can suppress fire risk in fuel-limited regions, mirroring the humped relationship seen for NDVI and population density. In our model, it always co-occurs with low humidity/precipitation.
& \cite{hantson2016status, bowman2020vegetation} \\ \hline

\end{tabular}
\caption{Details on the categories used for the fire-specific SAE-Xplainers semantic analysis in Figure 6 and their scientific meaning.} 
\label{tab:cat-desc}
\end{table*}

\begin{figure*}[tp]
    \centering
    \begin{subfigure}{0.48\textwidth}
        \centering
        \includegraphics[width=\linewidth]{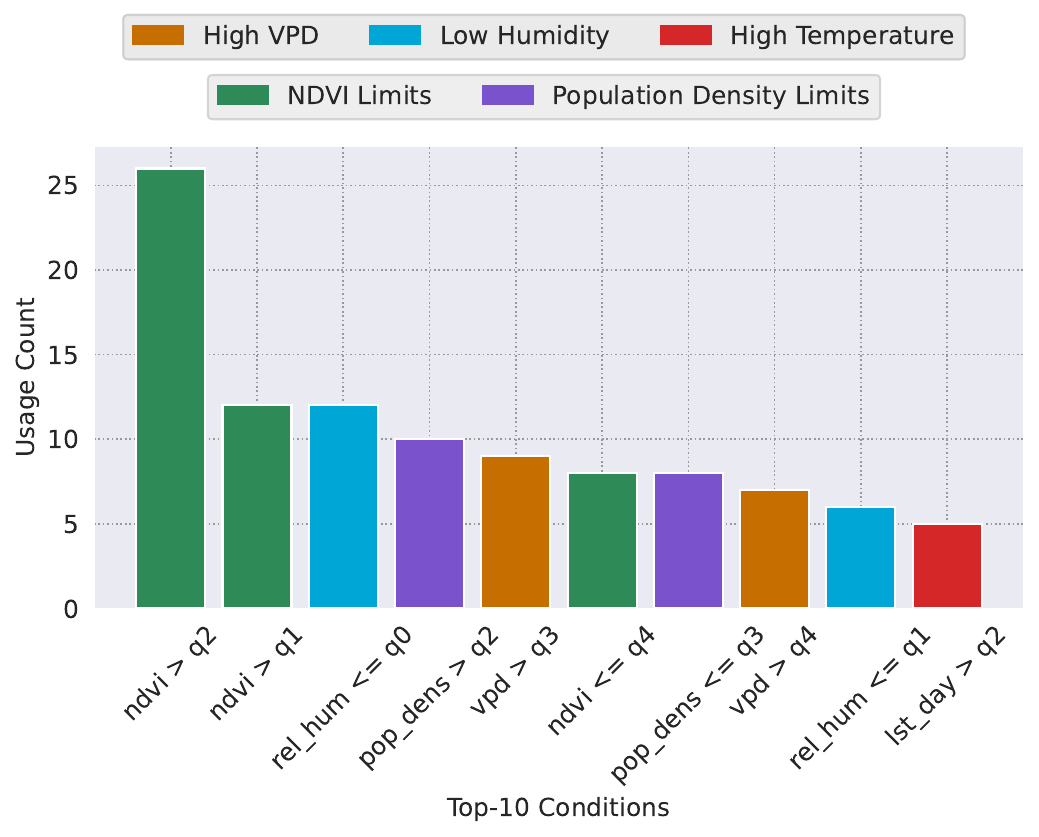}
        \caption{Histogram of most used conditions.}
        \label{fig:cond}
    \end{subfigure}
    \hfill
    \begin{subfigure}{0.48\textwidth}
        \centering
        \includegraphics[width=\linewidth]{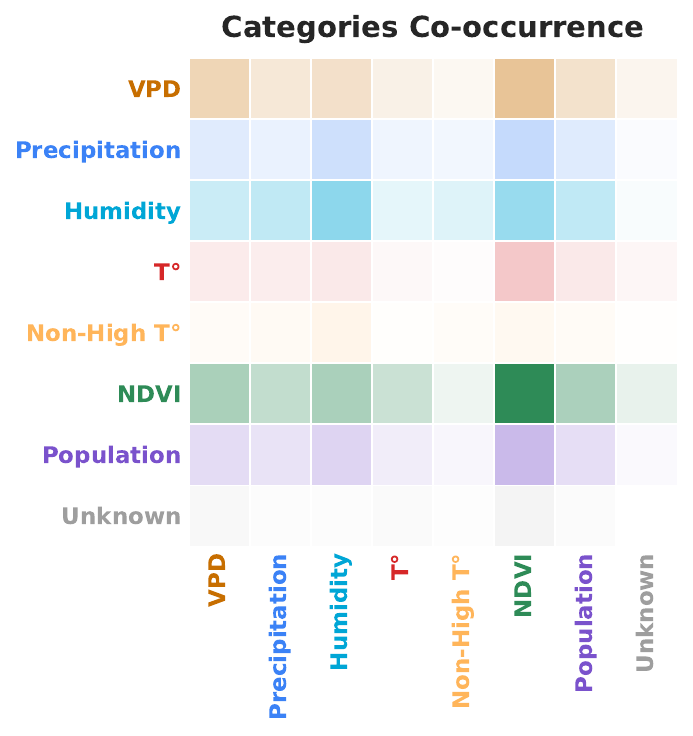}
        \caption{Co-occurrence matrix of high fire risk categories.}
        \label{fig:cat-oc}
    \end{subfigure}
    \caption{\textbf{(a):} Top 10 most used conditions (ignoring latitude and longitude) across the fire-specific SAE-Xplainers. The 20\% quantiles are denoted: q0, q1, q2, q3, and q4, and so $\text{ndvi} > \text{q2}$ means that NDVI is strictly superior to a threshold in the third 20\% quantile. \textbf{(b):} Co-occurrence matrix of the categories assigned to top-50 GeoTopK features. This matrix only records co-occurrence from distinct GeoTopK features, as they can be assigned to multiple categories.}
    \label{fig:cond-cooc}
\end{figure*}

\begin{figure*}[tp]
  \centering
  \includegraphics[width=\linewidth]{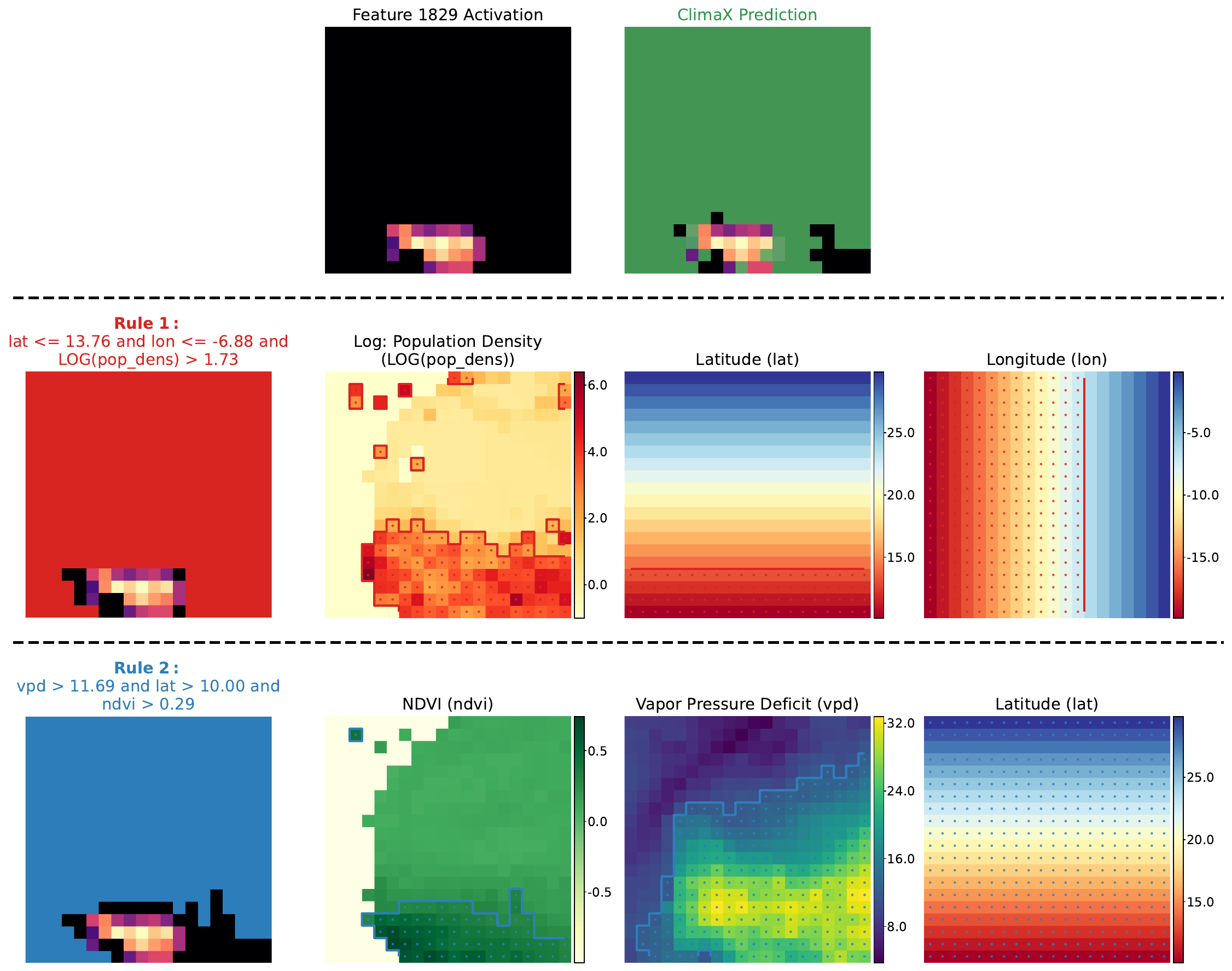}
   \caption{Sample-level feature interpretability for a ClimaX feature fine-tuned on SeasFire. \textbf{(Row 1):} Feature activations and model prediction masks (\textit{no fires} regions are masked). \textbf{(Rows 2--3):} SAE-Xplainer rule visualizations with binary masks (unsatisfied rule regions are masked) and input variables highlighting the rule support.}
    \label{fig:samp-fire-1}
\end{figure*}

\begin{figure*}[tp]
  \centering
  \includegraphics[width=\linewidth]{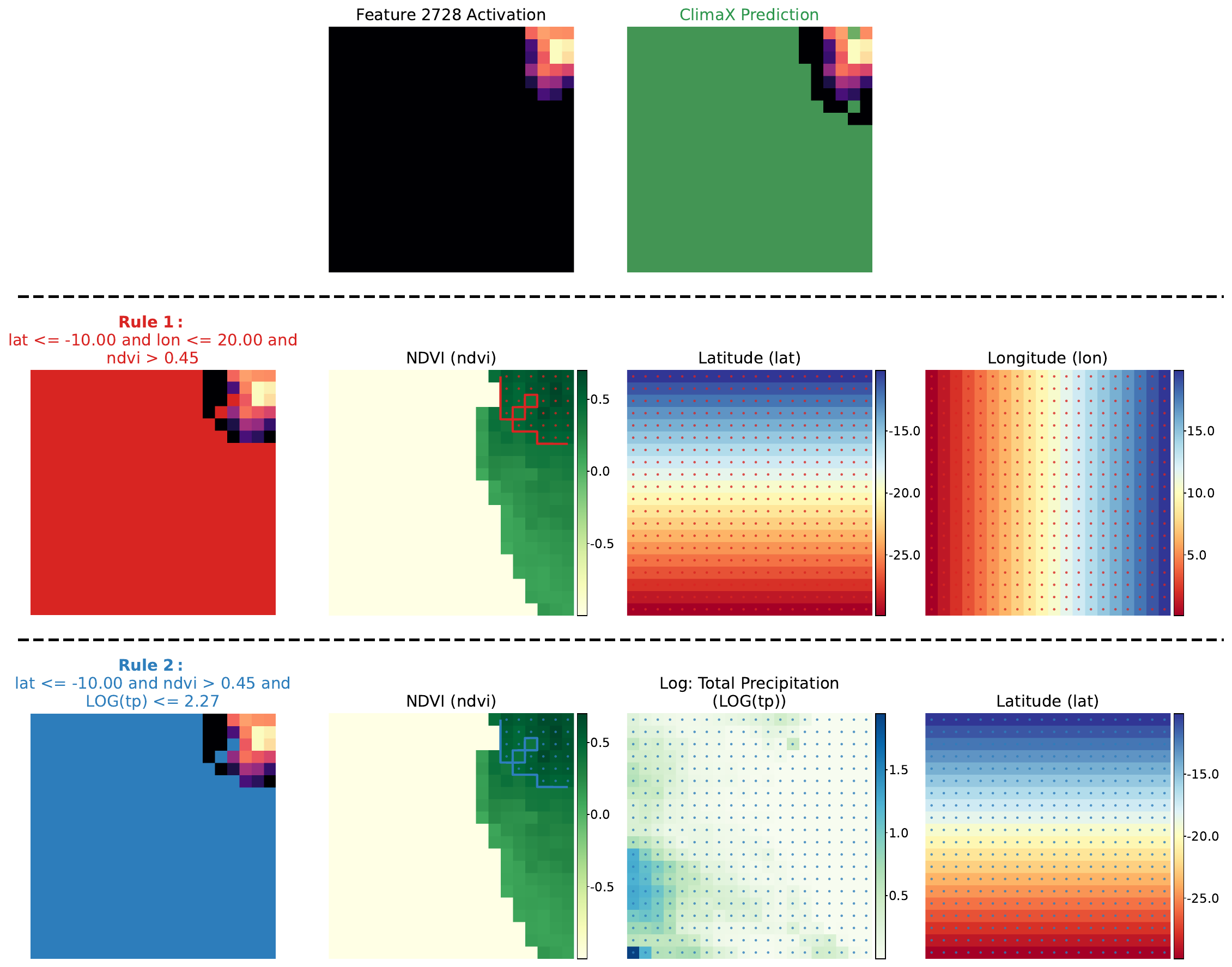}
   \caption{Sample-level feature interpretability for a ClimaX feature fine-tuned on SeasFire. \textbf{(Row 1):} Feature activations and model prediction masks (\textit{No Fire} regions are masked). \textbf{(Rows 2--3):} SAE-Xplainer rule visualizations with binary masks (unsatisfied rule regions are masked) and input variables highlighting the rule support.}
  \label{fig:samp-fire-2}
\end{figure*}

\begin{figure*}[tp]
  \centering
  \includegraphics[width=\linewidth]{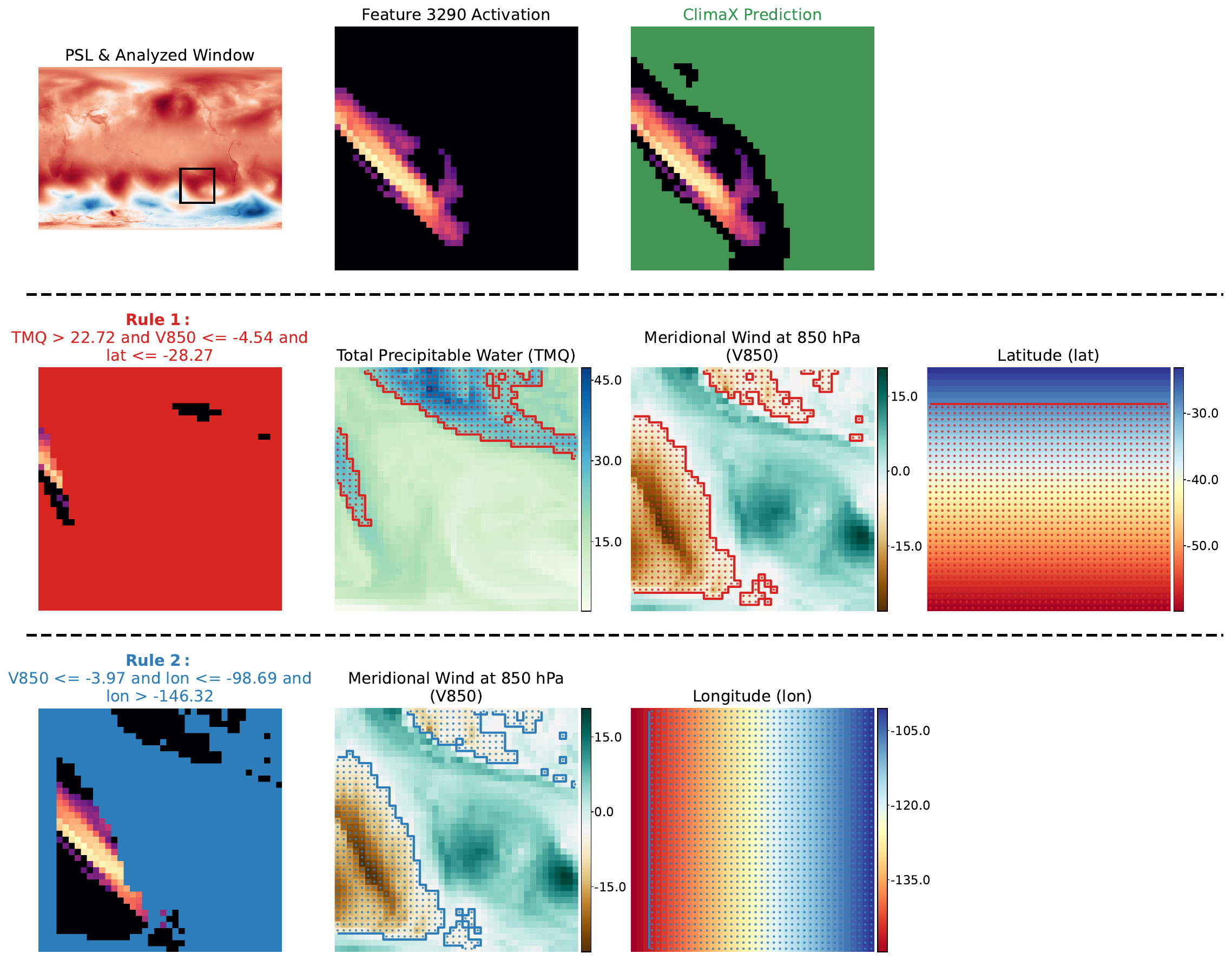}
   \caption{Sample-level feature interpretability for a ClimaX feature fine-tuned on ClimateNet and linked to \textit{Atmospheric Rivers}. \textbf{(Row 1):} Feature activations, model prediction masks (\textit{No Event} regions are masked), and high feature activation window of analysis. \textbf{(Rows 2--3):} SAE-Xplainer rule visualizations with binary masks (unsatisfied rule regions are masked) and input variables highlighting the rule support.}
  \label{fig:samp-ar-1}
\end{figure*}

\begin{figure*}[tp]
  \centering
  \includegraphics[width=\linewidth]{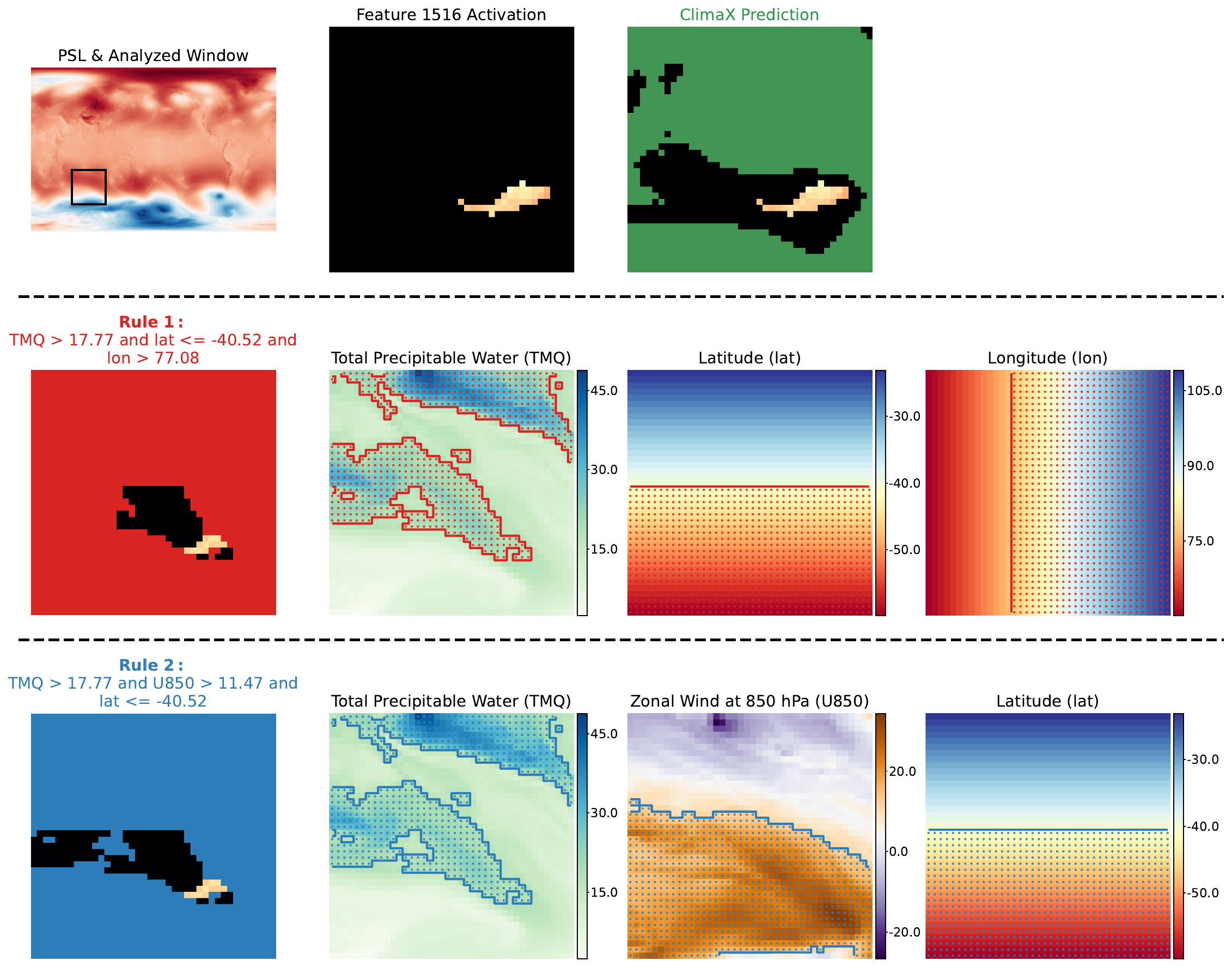}
    \caption{Sample-level feature interpretability for a ClimaX feature fine-tuned on ClimateNet and linked to \textit{Atmospheric Rivers}. \textbf{(Row 1):} Feature activations, model prediction masks (\textit{No Event} regions are masked), and high feature activation window of analysis. \textbf{(Rows 2--3):} SAE-Xplainer rule visualizations with binary masks (unsatisfied rule regions are masked) and input variables highlighting the rule support.}
  \label{fig:samp-ar-2}
\end{figure*}

\begin{figure*}[tp]
  \centering
  \includegraphics[width=\linewidth]{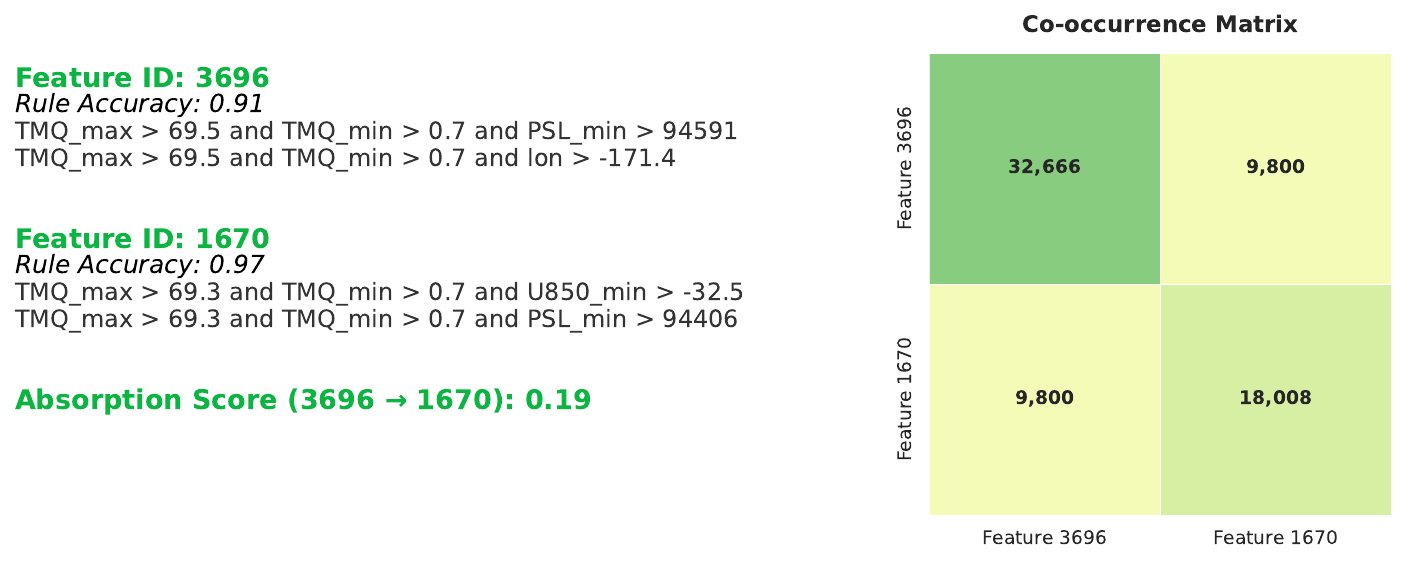}
   \caption{\textbf{(Left):} Example of identified feature group with absorption for ClimateNet. \textbf{(Right):} Co-occurrence matrix for those features computed over the whole dataset.}
  \label{fig:group}
\end{figure*}

\begin{figure*}[tp]
  \centering
  \includegraphics[width=\linewidth]{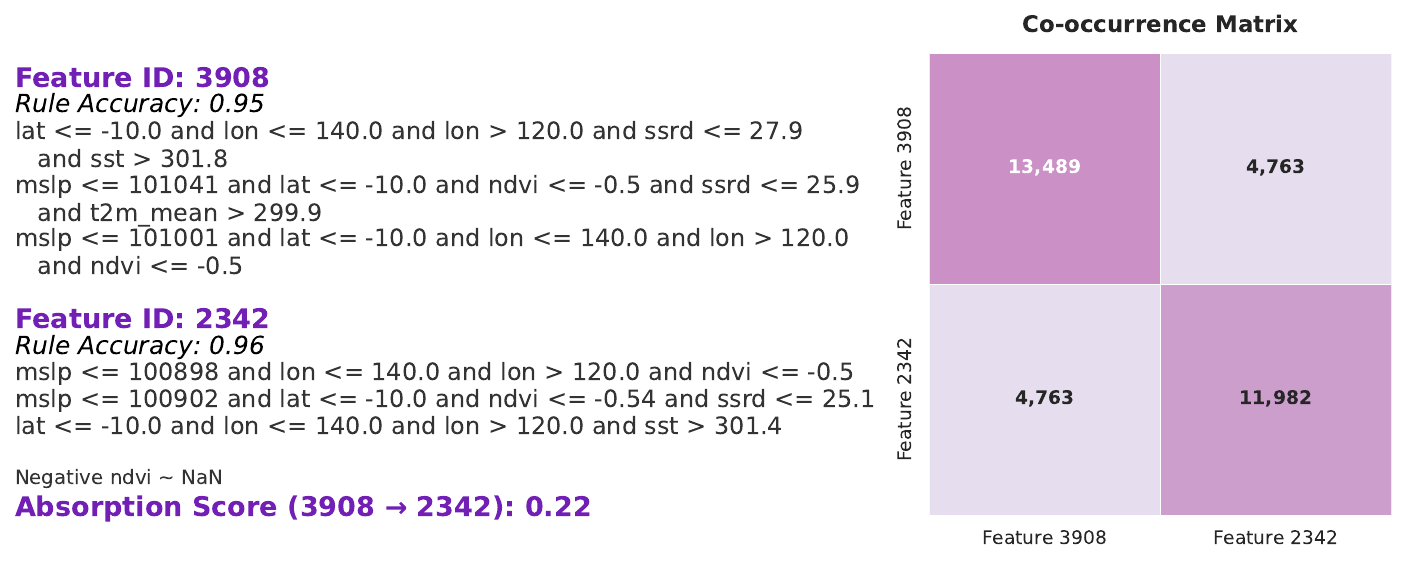}
   \caption{\textbf{(Left):} Example of identified feature group with absorption for SeasFire. \textbf{(Right):} Co-occurrence matrix for those features computed over the whole dataset.}
  \label{fig:group-2}
\end{figure*}

In Table~\ref{tab:cat-desc}, we provide the scientific descriptions with references from the wildfire literature of the categories used in the semantic structure analysis from Figure 6. This shows that the topics that emerge from the fire-specific SAE-Xplainers interpretations are deeply rooted in the wildfire literature and represent key global drivers of fire risk. In Figure~\ref{fig:cond}, we plot the most used rule conditions across our top-50 rule sets and observe that they can all be assigned to one of the categories from Table~\ref{tab:cat-desc}. We also analyze in Figure~\ref{fig:cat-oc} how the GeoTopK features specific to each category co-occur at the patch level on the SeasFire test set. For instance, the features assigned to the \textcolor{NDVIColor}{\textbf{NDVI Limits}} category, a descriptor of the local fuel load, co-occur with fire weather categories such as \textcolor{HighVPD}{\textbf{High VPD}} or \textcolor{LowHumidity}{\textbf{Low Humidity}} that indicate fuel and soil aridity. Lastly, we list in Table~\ref{tab:unk} the SAE-Xplainers that are mapped to the \textcolor{UNKColor}{\textbf{Unknown}} with the corresponding unidentified condition. 

\begin{table*}[tp]
\small
\setlength{\tabcolsep}{1.2mm}
\begin{tabular}{cccll}
\hline
\textbf{GeoTopK Feature} &
  \textbf{Eval. Accuracy} &
  \textbf{Complexity} &
  \textbf{Rule Set} &
  \textbf{Conditions} \\ \hline
1160 &
  0.92 &
  Mid &
  \begin{tabular}[c]{@{}l@{}}mslp <= 101174.00 and lat <= 9.36 and lat > 7.28 and ndvi > 0.31\\ mslp <= 101140.83 and lat <= 9.36 and lat > 7.28 and lon > 22.72\\ mslp <= 101157.69 and lat > 3.12 and lon > 26.88 and ndvi > 0.30\end{tabular} &
  \begin{tabular}[c]{@{}l@{}}\textbf{mslp <= q1}\\ ndvi > q2\end{tabular} \\ \hline
1222 &
  0.99 &
  Low &
  \begin{tabular}[c]{@{}l@{}}lat <= 10.00 and lat > 7.28 and lon <= 0.00\\ lat <= 10.00 and lon <= 0.00 and swvl1 > 0.08\\ \text{lst\_day} > 299.44 and lat <= 10.00 and lat > 7.28\end{tabular} &
  \begin{tabular}[c]{@{}l@{}}\textbf{swvl1 > q0}\\ \text{lst\_day} > q2\end{tabular} \\ \hline
1547 &
  0.96 &
  Low &
  \begin{tabular}[c]{@{}l@{}}lat <= -10.00 and lat > -11.68 and swvl1 > 0.08\\ lat <= -10.00 and lat > -11.68 and sst <= 116.95\\ lat <= -10.00 and lat > -11.68 and \text{pop\_dens} > -0.16\end{tabular} &
  \begin{tabular}[c]{@{}l@{}}\textbf{swvl1 > q0}\\ sst <= q0\\ \text{pop\_dens} > NaN\end{tabular} \\ \hline
944 &
  0.93 &
  Low &
  \begin{tabular}[c]{@{}l@{}}lat <= -4.16 and lat > -8.32 and \text{pop\_dens} > 0.40\\ lat <= -4.16 and lat > -8.32 and sst <= 74.93\\ lon <= 15.20 and lon > 13.12\end{tabular} &
  \begin{tabular}[c]{@{}l@{}}\text{pop\_dens} > q1\\ \textbf{sst <= q0}\end{tabular} \\ \hline
\end{tabular}
\caption{Description of Unknown SAE-Xplainer with in \textbf{bold} the rule condition to further analyze.}
\label{tab:unk}
\end{table*}

\end{document}